\documentclass[10pt,twocolumn,letterpaper]{article}

\usepackage[pagenumbers]{cvpr} %

\usepackage{makecell}
\usepackage[table, dvipsnames]{xcolor}
\usepackage{multirow}
\usepackage{arydshln}
\usepackage{gradient-text}
\usepackage[most]{tcolorbox}
\usepackage{adjustbox}
\usepackage{microtype}
\usepackage{comment}
\usepackage{amsmath}
\usepackage{listings}
\usepackage{newfloat}
\DeclareFloatingEnvironment[name=Listing]{listing}
\usepackage{tikz}
\usepackage{wrapfig}
\usetikzlibrary{positioning}
\renewcommand{\paragraph}[1]{\smallskip \noindent\textbf{#1}}
\newcommand{\methodname}{MemCoach\xspace}
\newcommand{\benchmarkname}{MemBench\xspace}
\newcommand{\supmat}{\textit{Supp.~Mat.\xspace}}

\newcommand{\modelQwen}{\textsc{Qwen2.5VL }~\citep{bai2025qwen25vltechnicalreport}}
\newcommand{\modelInternVL}{\textsc{InternVL3.5 }~\citep{wang2025internvl3}}
\newcommand{\modelIdefics}{\textsc{Idefics3 }~\citep{laurencconbuilding}}
\newcommand{\modelLLaVAOV}{\textsc{LLaVA-OV }~\citep{an2025llava}}
\newcommand{\modelGPTmini}{\textsc{GPT-5 Mini}~\citep{OpenAI_2025_GPT5_System_Card}}

\usepackage[most]{tcolorbox}
\definecolor{block-gray}{gray}{0.96}

\definecolor{zsblue}{RGB}{135, 206, 235}
\definecolor{ftpink}{RGB}{235, 135, 207}
\definecolor{methodgreen}{RGB}{213,232,212}
\definecolor{oraclered}{RGB}{248,206,204}
\definecolor{specializedyellow}{RGB}{240,174,14}

\definecolor{refs-blue}{HTML}{367dbd}

\definecolor{positiveblue}{HTML}{55C2F8}
\definecolor{negativered}{HTML}{f1312a}
\definecolor{inferencesteeringyellow}{HTML}{ffe776}
\definecolor{steeringextractionblue}{HTML}{a6caec}
\definecolor{contrastivedatapink}{HTML}{f1ceed}

\newcommand{\gradienttext}[1]{\gradientRGB{#1}{241, 50, 42}{85, 194, 248}}
\newcommand{\range}[2]{#1,\dots, #2}

\newcommand{\gain}[1]{\footnotesize{\textcolor{ForestGreen}{(+#1\%)}}}
\newcommand{\drop}[1]{\footnotesize{\textcolor{BrickRed}{(-#1\%)}}}
\newcommand{\perplexitygain}[1]{\footnotesize{\textcolor{ForestGreen}{(-#1\%)}}}
\newcommand{\perplexitydrop}[1]{\footnotesize{\textcolor{BrickRed}{(+#1\%)}}}
\newcommand{\ourslong}{Memorability Feedback\xspace}
\newcommand{\ours}{MemFeed\xspace}

\usepackage{dsfont}
\usepackage{subcaption}
\DeclareCaptionLabelFormat{andfigure}{#1~#2~\&~\figurename~\thefigure}

\newcommand{\ccc}[1]{%
  \ifnum\fpeval{#1 < -20}=1
    \cellcolor{DrawioRed!150}%
  \else\ifnum\fpeval{#1 < -10}=1
    \cellcolor{DrawioRed!100}%
  \else\ifnum\fpeval{#1 < 0}=1
    \cellcolor{DrawioRed!50}%
  \else\ifnum\fpeval{#1 > 20}=1
    \cellcolor{DrawioGreen!150}%
  \else\ifnum\fpeval{#1 > 10}=1
    \cellcolor{DrawioGreen!100}%
  \else\ifnum\fpeval{#1 > 0}=1
    \cellcolor{DrawioGreen!50}%
  \else
    \cellcolor{white}%
  \fi\fi\fi\fi\fi\fi
  #1%
}

\newcommand{\inlineColorbox}[2]{\begingroup\setlength{\fboxsep}{1pt}\colorbox{#1}{\hspace*{2pt}\vphantom{Ay}#2\hspace*{2pt}}\endgroup}

\definecolor{memregressor}{HTML}{F3E8FF}

\newcommand{\gpt}{\textsc{GPT-5-Mini}}
\newcommand{\llava}{\textsc{LLaVA-OV}}
\newcommand{\idefics}{\textsc{Idefics3}}
\newcommand{\qwen}{\textsc{Qwen2.5VL}}
\newcommand{\internvl}{\textsc{InternVL3.5}}

\newtcolorbox{usermsg}{
  colback=blue!10,
  colframe=blue!60,
  width=\linewidth,
  arc=4pt,
  outer arc=4pt,
  boxrule=0.4pt,
  left=2pt,
  right=2pt,
  top=4pt,
  bottom=4pt,
  enhanced,
}

\newtcolorbox{systemmsg}{
  colback=gray!10,
  colframe=gray!60,
  width=\linewidth,
  arc=4pt,
  boxrule=0.4pt,
  left=2pt,
  right=2pt,
  top=4pt,
  bottom=4pt,
  enhanced,
}

\newtcolorbox{botmsg}{
  colback=gray!10,
  colframe=gray!60,
  width=\linewidth,
  arc=4pt,
  outer arc=4pt,
  boxrule=0.4pt,
  left=2pt,
  right=2pt,
  top=4pt,
  bottom=4pt,
  enhanced,
}

\newtcolorbox{qwencaption}{
  colback=green!10,
  colframe=green!50!black,
  width=\linewidth,
  arc=4pt,
  outer arc=4pt,
  boxrule=0.4pt,
  left=2pt,
  right=2pt,
  top=4pt,
  bottom=4pt,
  enhanced,
}

\definecolor{stringcolor}{RGB}{186,33,33}

\definecolor{cvprblue}{rgb}{0.21,0.49,0.74}
\usepackage[pagebackref,breaklinks,colorlinks,allcolors=cvprblue]{hyperref}

\def\paperID{****} %
\def\confName{CVPR}
\def\confYear{2026}

\title{How to Take a Memorable Picture? \\Empowering Users with Actionable Feedback}

\author{
{Francesco Laiti\textsuperscript{1,2,3}
\quad
Davide Talon\textsuperscript{4}
\quad
Jacopo Staiano\textsuperscript{1}
\quad
Elisa Ricci\textsuperscript{1,4}}\\[-8mm]
\and
\normalsize
{\textsuperscript{1}University of Trento\quad \textsuperscript{2}University of Pisa\quad \textsuperscript{3}Travelbrain srl\quad
\textsuperscript{4}Fondazione Bruno Kessler}\\[2mm]
{\href{https://laitifranz.github.io/MemCoach/}{\normalsize \texttt{laitifranz.github.io/MemCoach}}}
}

\begin{document}
\maketitle

\begin{abstract}
Image memorability, \textit{i.e.}, how likely an image is to be remembered, has traditionally been %
studied in computer vision either as a passive prediction task, with models regressing a scalar score, or with generative methods altering the visual input to boost the image likelihood of being remembered. %
Yet, none of these paradigms supports users at capture time, when the crucial question is how to improve a photo memorability. %
We introduce the task of %
\textbf{Mem}orability \textbf{Feed}back (\textbf{MemFeed}), %
where an automated model should provide actionable, human-interpretable guidance to users with the goal to enhance an image future recall. We also present \textbf{\methodname}, the first approach designed to provide concrete suggestions in natural language for memorability improvement %
(e.g., “emphasize facial expression,” “bring the subject forward"). %
Our method, based on Multimodal Large Language Models (MLLMs), is training-free and employs a teacher-student steering strategy, aligning the model internal activations toward more memorable patterns learned from a teacher model progressing along least-to-most memorable samples. To enable systematic evaluation on this novel task, we further introduce \textbf{\benchmarkname}, a new benchmark %
featuring sequence-aligned photoshoots with annotated memorability scores. %
Our experiments, considering multiple MLLMs, demonstrate the effectiveness of \methodname, showing consistently improved performance over several zero-shot models. The results indicate that memorability can not only be predicted but also taught and instructed, shifting the focus from mere prediction to actionable feedback for human creators.
\end{abstract}
    
\section{Introduction} \label{sec:intro}
\begin{figure}[t]
    \centering
    \includegraphics[width=0.91\linewidth]{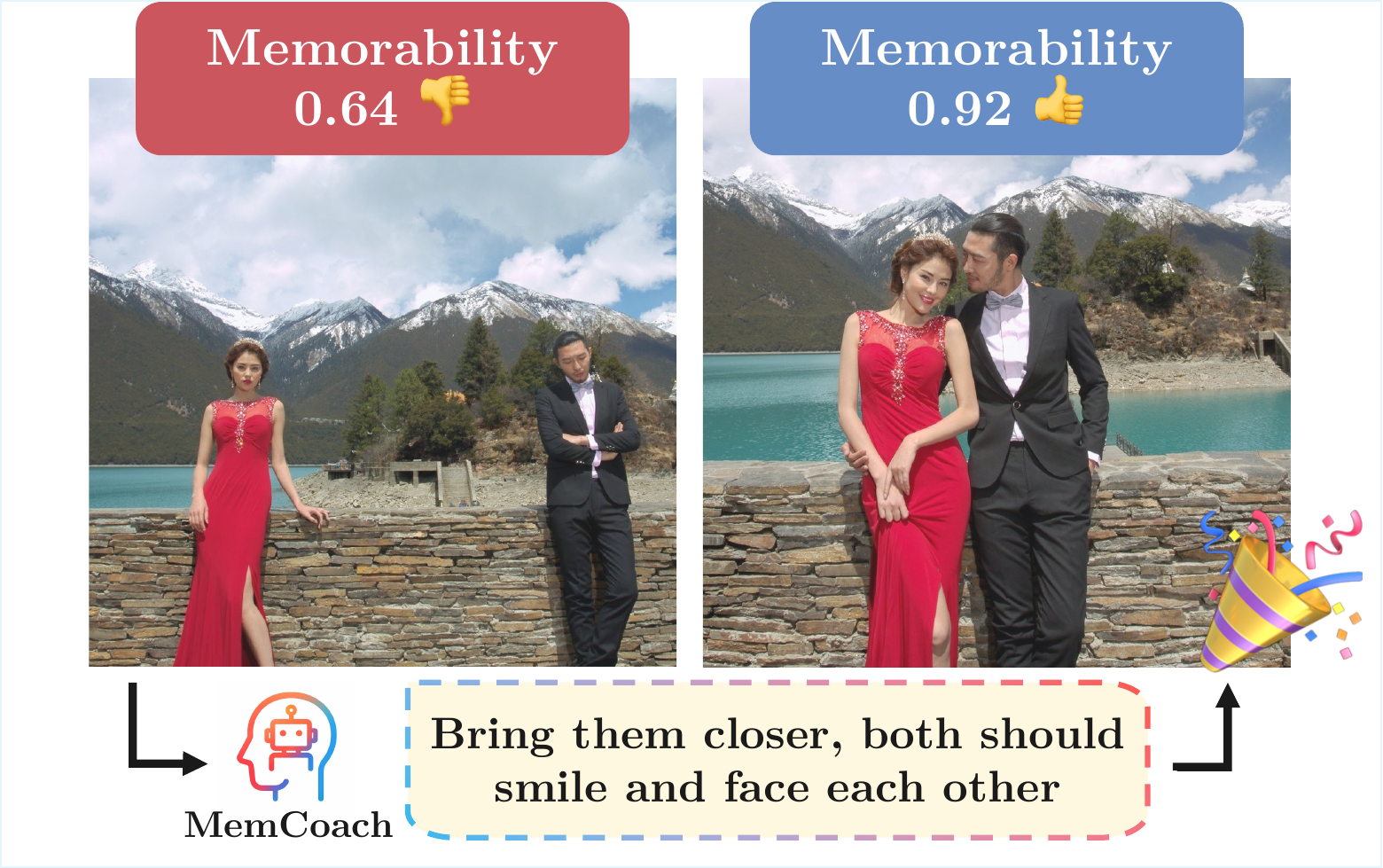}
    \caption{Given an input photo, \textit{memorability feedback} aims to generate \textit{natural-language suggestions} to guide users toward a more memorable shot. \methodname provides memorability-aware feedback, effectively \textit{assisting users} to capture memorable images.}
    \label{fig:teaser}
\end{figure}
Memorability, \ie{}, the likelihood that an image will be remembered by human observers, is an intrinsic property of a picture that can be predicted from visual content alone~\citep{isola2011makes,khosla2012image, bylinskii2015intrinsic, isola2011understanding}. Previous research has largely focused on measuring this property by introducing prediction models trained to regress a scalar memorability score from images~\citep{khosla2015understanding, squalli2018deep, fajtl2018amnet} and explaining what makes an image memorable~\citep{isola2011makes, isola2011understanding, isola2013makes}. These works identified key intrinsic factors such as the presence of people~\citep{dubey2015makes}, indoor scenes~\citep{bylinskii2021memorability}, or emotional expressions~\citep{bylinskii2021memorability}, rather than objects and panoramic views~\citep{isola2011makes}, as well as extrinsic ones, including context and the observer~\citep{bylinskii2015intrinsic}. More recent generative approaches have attempted to manipulate memorability, leveraging editing models to automatically enhance an image’s likelihood of being remembered~\citep{goetschalckx2019ganalyze, siarohin2017make}.
However, these paradigms are inherently passive and opaque: prediction models merely report how memorable an image is, while generative models directly alter the image, losing control on the changes. In contrast, when taking a picture, humans seek actionable feedback: \textit{``What should I change in this shot to make it more memorable?''}, rather than a numerical score or an automated edit, especially considering that, as humans, we generally fail to judge what is memorable~\cite{isola2013makes}. Similarly, in the context of computational photography, scoring models have been developed to  assess the quality of images \cite{wu2022interpretableaestheticanalysismodel, li2020personality} or to produce free-form critiques that are often verbose and difficult to operationalize as constructive feedback~\cite{Qi_2025_CVPR}.  

To address this gap, %
we introduce \textbf{Mem}orability \textbf{Feed}back (\textbf{\ours}), the task of providing users with \textit{actionable} and \textit{interpretable feedback} to improve image memorability. Instead of predicting or editing, an \textit{automated model} is used to provide \textit{guiding feedback}: given a user’s image, it generates natural-language suggestions describing concrete compositional or semantic changes that could increase memorability (\textit{e.g.}, \textit{``bring the subjects closer''}, \textit{``the subjects should smile and face each other''}), effectively verbalising how to improve the shot in terms of memorability (see \cref{fig:teaser}). Leveraging the reasoning and vision-language capabilities of {Multimodal Large Language Models (MLLMs)}, we propose \textbf{\methodname}, a novel approach that bridges perceptual memorability research and photographic assistance.
\methodname employs a training-free steering strategy that redirects MLLM activations toward memorability-aware feedback, \ie, suggestions enhancing memorability, as distilled from a teacher model indicating how to transition from less to more memorable images across multiple views of the same scene. This contrasts with the model’s default neutral feedback, which lacks memorability awareness.

To evaluate methods on the novel task of \ours, we introduce \textbf{\benchmarkname}, a new benchmark based on the PPR10K~\citep{liang2021ppr10k} dataset. It includes multiple images from the same photoshoot, each annotated with its memorability score. 
The proposed evaluation metrics are based on the quality of the model feedback, 
\ie, the memorability difference between the image the model is currently observing and the one after feedback implementation (as estimated by an editing model), as well as their perplexity on ground-truth effective feedback. Across four open-source MLLMs, our experiments show that MemCoach consistently enhances performance over standard zero-shot models. %

\textbf{Our contribution} is three-fold:
\begin{itemize}
    \item We investigate and formalize the task of \ourslong, where a model should provide human-understandable and actionable feedback to a user on how to make a shoot more memorable. To the best of our knowledge, this problem has not been previously studied.   
    \item We introduce \benchmarkname, a benchmark for memorability feedback training and evaluation.
    \item We present \methodname, a novel training-free method leveraging a teacher-student strategy and activation steering to inject memorability information for useful guidance. Our results show that \methodname can be effectively applied to multiple MLLMs.
\end{itemize}

\section{Related Work}
\label{ch.Related}

\paragraph{Memorability.}
Memorability refers to the probability that an observer will recall an image or a video after a quick view of it~\cite{isola2011makes, isola2011understanding, isola2013makes, bylinskii2015intrinsic, khosla2015understanding}. Early research \cite{isola2011makes, isola2011understanding} revealed that this is not a subjective phenomenon: memorability is a quantifiable property of visual content that is stable across observers. This property holds for both images~\cite{goetschalckx2019memcat, khosla2015understanding, isola2011makes, newman2020multimodal, lahrache2022survey, perera2019image, zalcher2025dontjudgeclipunified} and videos~\cite{SavranKiziltepe2021, cohendet2019videomem, newman2020multimodal, si2024longtermadmemorabilityunderstanding, martin2025parameter, dumont2023modular, kumar2025seeing}. Thus, the community has focused on understanding the intrinsic property underlying memorable visual content, finding that semantics plays a crucial role with faces and animals~\citep{dubey2015makes}, things~\cite{dubey2015makes}, indoor~\citep{bylinskii2021memorability} or less cluttered scenes~\citep{goetschalckx2019ganalyze} and images conveying negative emotions~\citep{bylinskii2021memorability} having an increased memorability score. This is in stark contrast with the original belief that natural vistas and aesthetic beauty make an image memorable~\citep{isola2011understanding}. Furthermore, researchers have investigated the influence of extrinsic factors like visual context, eye movements and the role of the observer~\cite{bylinskii2015intrinsic}. Most similar to our work, \citep{goetschalckx2019ganalyze, khosla2013modifying, siarohin2017make, sidorov2019changing} leverage editing models for increasing the memorability of images at hand. In contrast, our goal is to develop models that provide users with natural language feedback on how to enhance an image’s memorability.%

\paragraph{Photographic feedback.}
Recent efforts~\cite{qi2025photographer,wu2024q, huang2024aesexpert} focus on curating photograph datasets annotated with professional critiques and aesthetic feedback. Models trained on these data can explain compositional strengths and weaknesses in natural language, providing users with critique-like feedback. However, they fall short on translating critique into concrete, actionable instructions that a user can execute at the moment of shooting.
Research on photographic guidance has largely focused on aesthetic scoring or rule-based feedback~\cite{wu2021tumera, kahlon2025portraid, s20030582} (\textit{e.g.}, rule of thirds) with overlays to assist the user. Similarly, works like \cite{10.1145/3313831.3376635, ma2019smarteye, Limoyo_2024} propose diversified views and adaptive composition grids to improve image quality. While effective for novice photographers, these approaches mainly offer static rule enforcement or post-hoc critique rather than adaptive, scene-specific coaching. Most recently, \cite{Google_2025_CameraCoach} highlights the increasing demand for interactive photographic guidance. However, such systems are proprietary, and a formalized framework, including publicly available benchmarking data and evaluation metrics, is still lacking in the literature.

\paragraph{MLLMs and steering.}
Starting from early approaches limited to learning a shared embedding space where visual and textual representations are aligned~\cite{radford2021learningtransferablevisualmodels, zhai2023sigmoidlosslanguageimage}, recent research efforts have focused on generative methods capable of coherent question-answering%
~\cite{li2023blip2bootstrappinglanguageimagepretraining, alayrac2022flamingovisuallanguagemodel, liu2023visual, zhu2024minigpt, huang2023language, bai2023qwen}. Under the linear activations hypothesis~\citep{park2023linear}, steering approaches~\citep{turner2023steering, zou2023representation, rimsky2024steering} show that model behaviour can be modified via linear displacements of model intermediate representations~\citep{turner2023steering, zou2023representation, rimsky2024steering}. Steering typically involves building contrasting sample sets differing in a target concept, computing a mean-difference vector, and adding or subtracting it from activations to control the concept at inference time~\citep{chen2025persona, facchiano2025videounlearninglowrankrefusal, shen2025lunar, talon2025seeing, rimsky2024steering, belitsky2025kv}. 
In contrast with these works, we design a teacher-student steering approach for actionable memorability feedback. To the best of our knowledge, \methodname is the first activation steering strategy for MLLMs applied to perceptual tasks.

\begin{figure*}[ht!]
    \centering
    \includegraphics[width=0.9\linewidth]{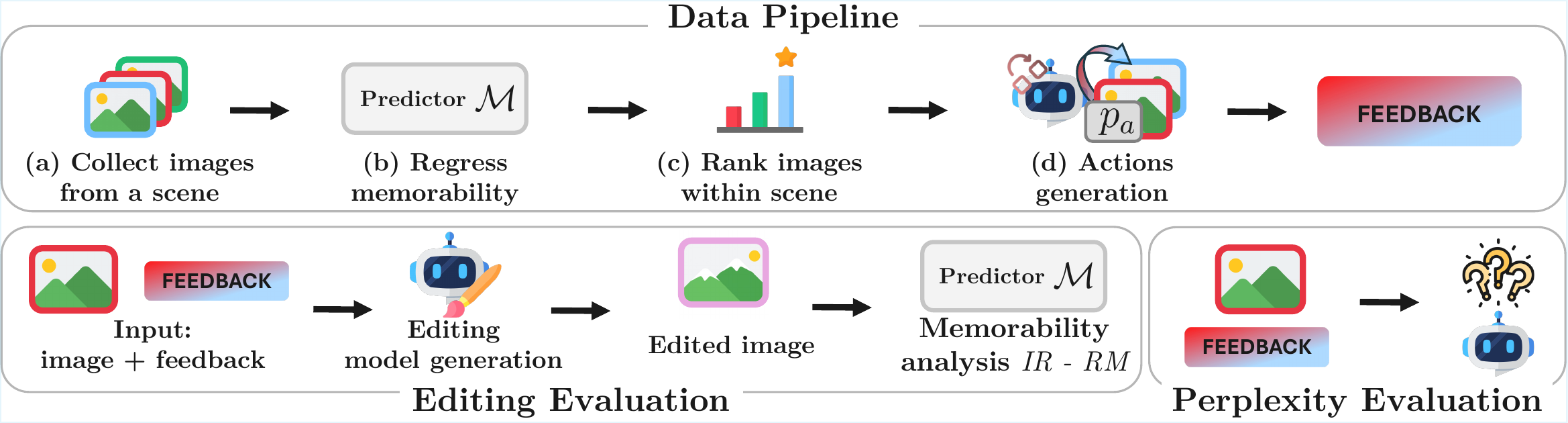}
    \caption{\textbf{Overview of \benchmarkname generation and evaluation}. \textbf{Top:} Data pipeline for constructing \benchmarkname, including scene grouping, memorability regression, image ranking, and generation of actionable memorability-aware feedback.
    \textbf{Bottom:} Evaluation pipeline assessing feedback quality through editing-based memorability improvement and perplexity scoring.}
    \label{fig:benchmark}
\end{figure*}

\section{Memorability Feedback}
\label{sec:benchmark}
In this section, we formally define the task of \ourslong and present \textbf{\benchmarkname{}}, a novel benchmark to test the effectiveness of models in providing actionable and human-interpretable guidance to take memorable images.

\subsection{Task Definition}
We frame the memorability feedback task as a transformation problem over visual content. Given a source image $x_{S}$ with an associated memorability score $m_{S} \in [0,1]$, the objective is to design an automated model capable of generating a natural language actionable feedback $a$ that, when implemented on $x_S$, would get to the destination image $x_{D}$, such that the resulting memorability score $m_{D}$ satisfies $m_{D} > m_{S}$ (\cref{fig:teaser}). Here, we assume $m_{D} = \mathcal{M}(x_{D})$ and $m_{S} = \mathcal{M}(x_{S})$ are estimated by a memorability prediction model $\mathcal{M}$. This task departs from conventional memorability prediction, as it requires models not only to assess the current memorability level, but to \emph{proactively} identify and verbalize actions capable of increasing it. The generated feedback must be both semantically grounded in the visual content and operationally feasible. In this formulation, success depends on the model’s ability to reason about image properties that influence human memory and to translate such reasoning into targeted and constructive guidance.

\subsection{\benchmarkname} \label{sec:membench}
We introduce \benchmarkname, a benchmark for memorability-aware feedback. \benchmarkname builds upon PPR10K \cite{liang2021ppr10k} by augmenting image pairs, of the same scene, with natural-language semantic action descriptions that specify how the visual content differs between a lower-memorability image and a higher-memorability counterpart. 
\begin{figure*}[t!]
    \centering
    \includegraphics[width=.8\linewidth]{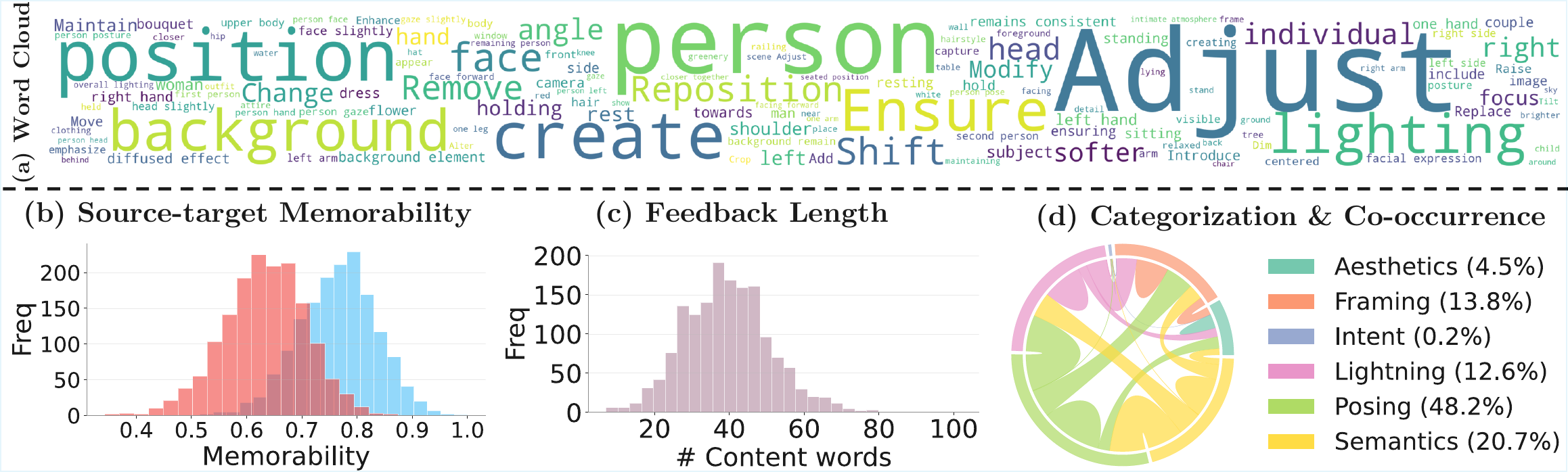}
    \caption{\textbf{\benchmarkname statistics.} Data analysis in terms of \textbf{(a)} most frequent words; \textbf{(b)} distribution of memorability scores for the least and most memorable images within each scene; \textbf{(c)} feedback length as measured by content words; and \textbf{(d)} categorization of atomic sub-actions, where the width of each chord indicates the frequency of co-occurrence between categories.}
    \label{fig:benchmark-stats}
\end{figure*}

\paragraph{Data pipeline.}
We built upon PPR10K~\citep{liang2021ppr10k}, a portrait photo retouching dataset with several different scenes. PPR10K offers multiple shoots per scene, where each taken photograph may differ both in subjects and composition as well as framing and lighting. 
A visualization of the data pipeline is depicted in ~\cref{fig:benchmark}. We start by collecting images from the same scene and group them together (\cref{fig:benchmark}-a). In a second step, for each image we evaluate its memorability by means of a predictor $\mathcal{M}$, a pre-trained regressor (\cref{fig:benchmark}-b) built upon CLIP \citep{radford2021learningtransferablevisualmodels} features and trained on publicly available memorability datasets \citep{khosla2015understanding, goetschalckx2019memcat, isola2011makes}, reaching state-of-the-art performance. See \supmat{} for predictor $\mathcal{M}$ details.
Once images are associated to the corresponding memorability score, photographs within the same scene are ranked and pairs $\left(x_S, x_D\right)$ are constructed from less to more memorable images (\cref{fig:benchmark}-c). %

\paragraph{Extracting actionable memorability feedback.}
For each image pair $(x_S, x_D)$, %
we prompt a captioning model $\psi$ that allows for interleaved images, to describe the feedback $a$ necessary to transform the source into the destination image (\cref{fig:benchmark}-d):
$a = \psi(x_S, x_D, p_a)$
where $p_a$ is the feedback elicitation prompt: ``\texttt{Determine the actions required to transform} $\langle x_S \rangle$ \texttt{into} $\langle x_D\rangle$''.
Contrary to computational photography adjustments focusing on post-hoc corrections (\textit{e.g.}, \textit{``make the image brighter''}), we focus on semantic actions that a user can take on-the-fly for a better shot, \textit{e.g.}, \textit{``Face each other''} (see \supmat{} for qualitative samples). We rely on \textsc{InternVL3.5 8B}~\citep{wang2025internvl3} as captioning  model.

\paragraph{Benchmark statistics.} \benchmarkname{} comprises approximately 10K images grouped into 1,570 scenes, with an average of 6.5 images per scene. The word cloud in \cref{fig:benchmark-stats}-a illustrates the most frequent terms appearing in the collected feedback. As shown, suggestions span a wide range of semantic categories, including references to body parts (\textit{e.g.}, ``\textit{hand}'', ``\textit{face}''), verbs (\textit{e.g.}, ``\textit{holding}'', ``\textit{remove}''), and photographic concepts (\textit{e.g.}, ``\textit{background}'', ``\textit{lighting}'').
Source images exhibit an average memorability score of 0.63, while the most memorable images within the same scene range between [0.51, 1.0], indicating some overlap between the two distributions (\cref{fig:benchmark-stats}-b). Feedback varies in length, ranging from 7 to 102 words (\cref{fig:benchmark-stats}-c). Finally, in \cref{fig:benchmark-stats}-d, we categorize atomic sub-actions in the feedback using \textsc{GPT-5-mini} as an automatic annotator and report their co-occurrence patterns (see \supmat{}). Most sub-actions relate to subject posing, followed by semantic adjustments, while co-occurrence statistics highlight strong correlations between framing and posing and the interplay between lighting and semantic changes. 

\paragraph{Evaluation protocol.} As we propose a novel task, we also introduce evaluation metrics for \ourslong (see Fig.\ref{fig:benchmark}-bottom), covering two main axes: \textit{real world effectiveness} and \textit{likelihood of memorable actions}. On the one hand, editing metrics probe the effectiveness of provided feedback by emulating real-world user behavior; we use \textsc{FLUX.1 Kontext}~\citep{labs2025flux1kontextflowmatching} as in-context image editing model $e(\cdot, \cdot)$ which applies the guidelines provided by the memorability feedback: starting from the source image $x_S$ and the feedback $a$, the destination image is obtained as edited output $\hat{x}_D = e(x_S, a)$. Hence, \textsc{Improvement~Ratio}~(IR) evaluates the fraction of time the edited image has larger memorability than the source one, \ie, $\text{IR} = \sum_{x_D} \mathds{1}\left[m_D \geq m_S\right]$, with $\mathds{1}[c]$ the indicator function evaluating the satisfaction of the $c$ condition. Instead, \textsc{Relative Memorability}~(RM) is defined as: $\text{RM} = (m_D-m_S)/m_S$, accounting for relative memorability improvements. In the \supmat{}, we report experiments with different editing models and a different memorability predictor. On the other hand, we evaluate the likelihood that a model provides improving memorability feedback by computing the \textsc{Perplexity} on ground truth memorability-aware feedback from the same captioning model. We use an 80-20 train/test scenes split, evaluating the feedback model only on scenes not seen during training.

\section{Method}
\label{sec:method}
Our goal is to design a model with the ability to provide actionable feedback or, in other words, suggestions that when applied to a user’s photo, can enhance its memorability. Given their capability to jointly interpret visual inputs and generate coherent textual descriptions, multimodal large language models are naturally well-suited for this task. However, when naively prompted, MLLMs lack a concrete understanding of what makes an image memorable (Sec.\ref{sec:method:mllm-mem}). In Sec.~\ref{sec:memolift} we hence describe how to enable a multimodal large language model to effectively perform \ours.

\subsection{MLLMs Lack Memorability Understanding}
\label{sec:method:mllm-mem}
Since even humans provide inconsistent judgments of memorability~\citep{isola2013makes}, we first investigate whether contemporary MLLMs are able to capture the underlying factors that make an image memorable. We conduct a preliminary study on the LaMem dataset~\citep{khosla2015understanding}, where each image is annotated with its memorability score. Specifically, we prompt recent MLLMs with a simple question, by asking whether a given image is memorable ``\texttt{Is this image memorable? Output only yes or no.}'' and interpret the likelihood of the \texttt{yes} token with respect to the \texttt{no} token as the predicted memorability score. 
Following prior works \citep{khosla2015understanding, goetschalckx2019memcat, isola2011makes}, we evaluate the results in terms of Spearman’s rank correlation~\citep{spearman1961proof} against the ground-truth scores. As shown in \cref{tab:mem_understanding}, despite extensive pretraining, MLLMs exhibit no correlation with human annotations, remaining far below the cross-annotator consistency upper bound. Consequently, they also fail to provide reliable or effective feedback for enhancing memorability (see Fig.~\ref{fig:mem_understanding}). Indeed, we observe a marginal IR improvement for all zero-shot models when prompted with $p_m=$ ``\texttt{Determine the actions required to improve the memorability of $\langle x_S \rangle$}'' compared to the \textsc{Editing baseline}, implemented by providing an empty-string instruction %
to the editing model $e(\cdot, \text{``\;''})$, leaving the image unaltered except for the model’s default bias. 

\begin{table}[t]
\begin{minipage}[c]{0.48\linewidth}
		\centering
		\resizebox{\textwidth}{!}{%
        \begin{tabular}{l c}
            \toprule
            \textbf{Model} & \makecell{\textbf{Spearman}\\\textbf{Rank} ($\uparrow$)} \\
            \midrule
            \rowcolor{oraclered!35} Inter-annotator$^*$ & 0.68 \\
            \qwen~\citep{bai2025qwen25vltechnicalreport} & -0.06 \\
            \internvl~\citep{wang2025internvl3} & -0.01 \\
            \idefics~\citep{laurencconbuilding} & -0.07 \\
            \llava~\citep{an2025llava} & 0.08 \\
            \bottomrule
        \end{tabular}
    }
	\end{minipage}\hfill
	\begin{minipage}[c]{0.48\linewidth}
  \centering
    \vspace{+1.2em}
      \includegraphics[width=\linewidth]{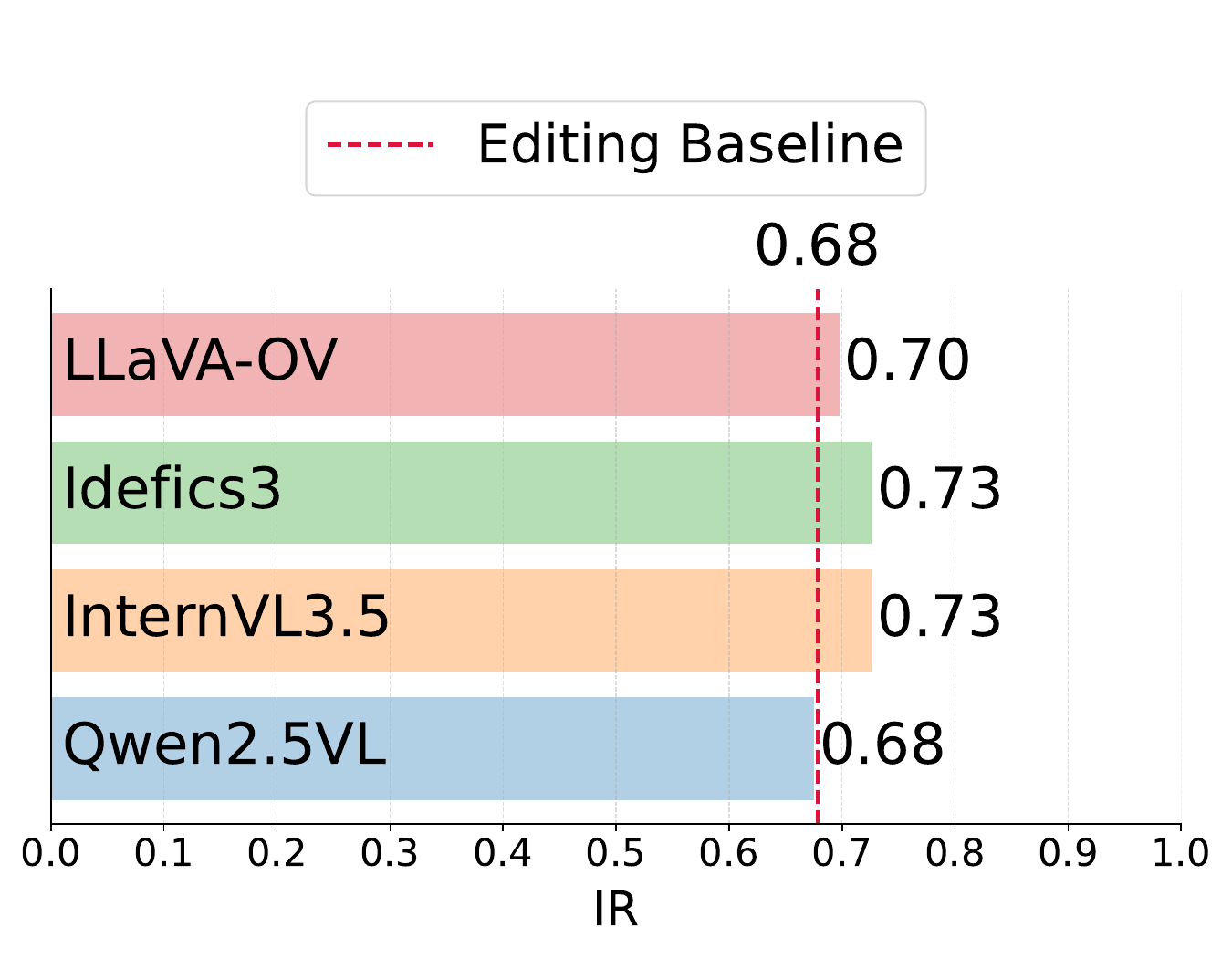}
      \refstepcounter{figure}
      \label{fig:mem_understanding}

\end{minipage}
\vspace{-1em}
 \captionsetup{labelformat=andfigure}
    \caption{\textbf{MLLMs lack memorability understanding.} \textbf{Left:} Memorability prediction performance in terms of Spearman’s Rank Correlation $ (\uparrow)$. %
    $(^*)$ is reported from~\citep{khosla2015understanding}. \textbf{Right:} Improvement ratio of zero-shot models with respect to the \textsc{Editing baseline}, marginal improvement is observed.}
    \label{tab:mem_understanding}
\end{table}

\subsection{\methodname}
\label{sec:memolift}
We introduce \methodname{}, a training-free approach to elicit memorability feedback in state-of-the-art MLLMs thanks to a novel knowledge-distillation activation steering strategy.

\begin{figure*}[t!]
    \centering
    \includegraphics[width=0.91\linewidth]{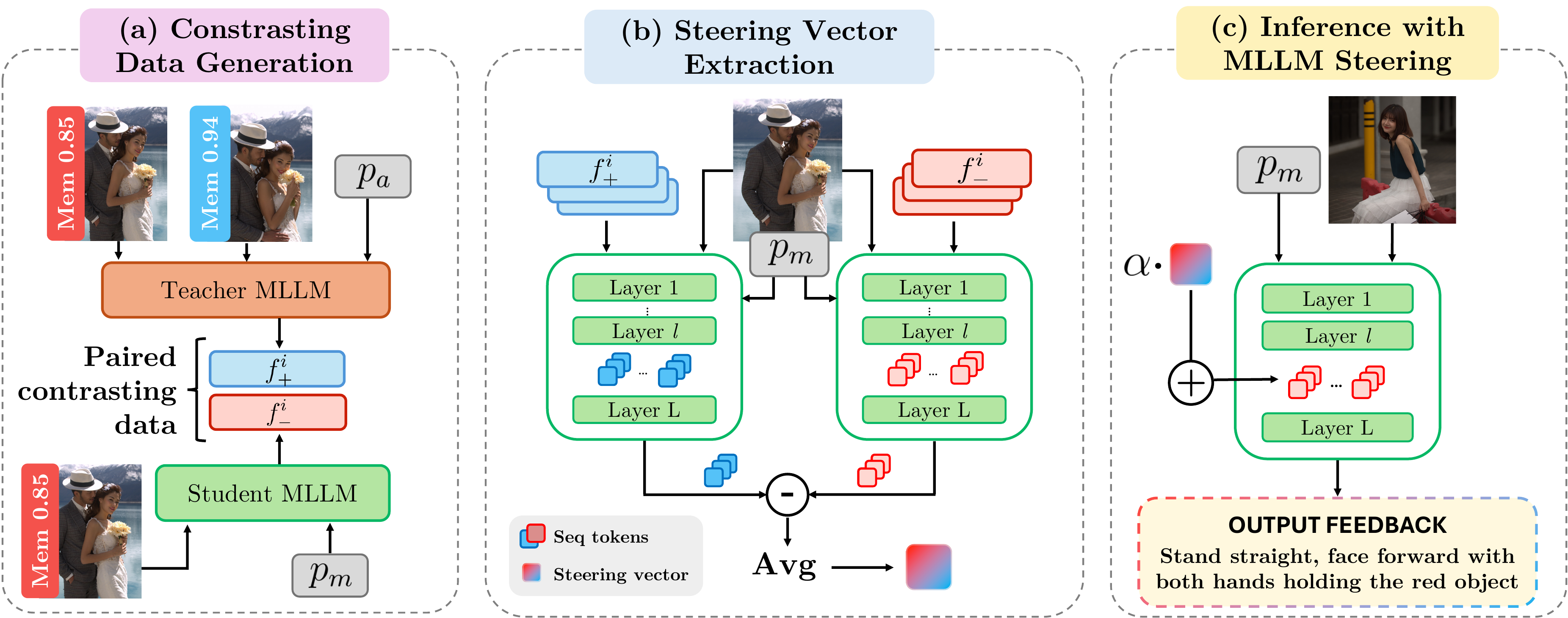}
    \caption{
        \textbf{Overview of the proposed method.}
        \inlineColorbox{contrastivedatapink!40}{\textbf{(a) Contrasting data generation}}: paired samples are built by coupling the \textcolor{positiveblue}{memorability-aware} guidance of a teacher MLLM with the \textcolor{negativered}{neutral} responses of a student MLLM on the same scene; 
        \inlineColorbox{steeringextractionblue!30}{\textbf{(b) Steering vector extraction}}: activation differences between memorability-aware and neutral feedback are averaged to obtain a \gradienttext{memorability steering vector} capturing the latent shift toward effective suggestions for memorability; 
        \inlineColorbox{inferencesteeringyellow!30}{\textbf{(c) Inference with MLLM steering}}: the student activations are shifted using the memorability steering vector to produce improved, memorability-oriented feedback without additional training.
    }    
    \label{fig:method}
\end{figure*}

\paragraph{Method overview.} \cref{fig:method} depicts our approach. In the initial \textit{contrasting data generation} step (\cref{fig:method}-left), \methodname leverages multiple images corresponding to the same scene to construct a paired dataset where the default behaviour of a \textit{student} MLLM asked to provide memorability feedback (\ie, neutral feedback) is compared to the one of a \textit{teacher} model generating actions that will transform the source image to a destination image that \textit{is known} to be more memorable (\ie, memorability-aware feedback). Then, the second \textit{steering vector extraction} step (\cref{fig:method}-center) extracts a memorability steering vector on student activations to capture the latent-space deviations introduced by memorability-aware feedback. Finally, at inference time (\cref{fig:method}-right), the \textit{MLLM steering} step uses such vector to shift the student model activations toward more effective suggestions.

\paragraph{\inlineColorbox{contrastivedatapink!40}{Contrasting data generation.}} 
We build paired memorability feedback samples based on the difference in memorability improvement that they induce.

 Formally, consider a dataset $\mathcal{D} = \{(\mathcal{X}^i)\}_{i=1}^N$ where for each scene $i$, the set of images $\mathcal{X}^i = \{x^i_1, \dots, x^i_M\}$ are captured within the same shooting session. 
 Our goal is to generate feedback pairs $(f^i_+, f^i_-)$. $f^i_+$ corresponds to the memorability-aware feedback provided by a teacher model which effectively describes how to get to more memorable images.~$f^i_-$ corresponds to the default student behavior when asked to suggest for improved memorability. To this end, we follow the data generation pipeline in~\cref{sec:benchmark}. Each image is evaluated for its memorability score with $\mathcal{M}$ and ranked accordingly.
 Consider $x^i_S$ as the least memorable image in $\mathcal{X}^i$, \textit{i.e.}, the source image we want to provide feedback on, and $x^i_D$ the most memorable image within the set, or, in other words, the desired output we would like to get with the provided feedback. Let the teacher model $\phi_\texttt{teach}$ be a MLLM that, when observing a pair of images $(x, x')$, elicits the corresponding actions to move from image $x$ to image $x'$ and $\phi_\texttt{stud}$ the student model we are interested to enable for effective feedback for memorability on an observed image $x$. On the one hand, we leverage the teacher model to extract \textcolor{positiveblue}{memorability-aware feedback} $f^i_+ = \phi_\texttt{teach}(x^i_S, x^i_D, p_a)$, with $p_a$ the feedback elicitation prompt in Sec. \ref{sec:benchmark}, yielding the actionable instructions on how to move from $x^i_S$ to $x^i_D$ and consequently, improve the current image memorability. On the other side, we collect $\phi_\texttt{stud}$ default \textcolor{negativered}{neutral feedback}, $f^i_- =\phi_\texttt{stud}(x^i_S, p_m)$, where $p_m$ is the memorability feedback prompt in \cref{sec:method:mllm-mem}. We construct paired contrasting data as:
\begin{equation}
\label{eq:contrasting-data}
        \mathcal{F}_+ = \{f^i_+\}_i, \quad \mathcal{F}_- = \{f^i_-\}_i,
\end{equation}
 with $i=\range{1}{N}$. In summary, paired data in ~\cref{eq:contrasting-data} captures the discrepancy between student-default and teacher-privileged memorability-aware feedback: for the same source image $x_S$ the privileged knowledge of the memorability target is opposed to the student’s default uninformed suggestions.

\paragraph{\inlineColorbox{steeringextractionblue!30}{Steering vector extraction.}} 
Starting from the available contrasting data, this step aims to characterize the student activation-space directions capturing the systematic shift between memorability-aware feedback and neutral one.

Despite both sets providing valid suggestions on the source image $x^i_S$, feedback in $\mathcal{F}_+$ improves memorability, whereas the ones in $\mathcal{F}_-$ have limited effect. Inspired by steering strategies~\citep{turner2023steering}, we therefore leverage their discrepancies to disentangle the factors that improve memorability.
To this end, we construct the input to the student model by using: 
\begin{align} \label{eq:input_steering_vects}
    \mathbf{f}^i_+ &= \{(x^i_S, p_m, f^i_+)\}_{f^i_+ \in \mathcal{F}_+}, \\
    \mathbf{f}^i_- &= \{(x^i_S, p_m, f^i_-)\}_{f^i_- \in \mathcal{F}_-},  \nonumber
\end{align}
where $f^i_+$ and $f^i_-$ are placed in the \texttt{assistant} turn of the chat template, paired with the same source image $x^i_S$ and prompt $p_m$, thus inducing different responses for identical inputs.
Then, we independently feed $\mathbf{f}^i_+$ and $\mathbf{f}^i_-$ to the student model %
to collect its activations on the two different types of feedback. Let define $h^{(l)}$ as the activation of $\phi_\texttt{stud}$ at layer $l=\range{1}{L}$, where $L$ is the number of its layers, and let $h^{i, (l)}_+$ and $h^{i, (l)}_-$ denote the aware and neutral feedback activations for the $i$-th sample at layer $l$. We extract the \gradienttext{memorability steering vector} $\mathbf{r}^{(l)}$ at layer $l$ as:
\begin{equation}
    \label{eq:steering-compute}
    \mathbf{r}^{(l)} = \frac{1}{N} \sum_{i=1}^N h_+^{i, (l)} -  h_-^{i, (l)}, 
\end{equation}
This vector characterizes the shift between memorability-aware and neutral feedback in the model activation space, acting as a \textit{distilled representation} of the teacher’s privileged knowledge, later used to steer the uninformed student toward more effective memorability guidance.

\paragraph{\inlineColorbox{inferencesteeringyellow!30}{Inference with MLLM steering.}}
At inference time, we aim to endow the student model with the 
capability to improve memorability, without relying on the teacher privileged information. Given a user-provided image $x$ and the memorability instruction prompt $p_m$, we first compute the student default activations $h^{(l)}$ and then steer the model by injecting the memorability steering vector $\mathbf{r}^{(l)}$ extracted in the previous step. Formally, the activations are shifted as:
\begin{equation}
\label{eq:steered-activation}
\tilde{h}^{(l)} = h^{(l)} + \alpha \cdot \mathbf{r}^{(l)},
\end{equation}
where $\alpha$ is a hyperparameter controlling the steering strength.
Intuitively, \cref{eq:steered-activation} shifts the model’s intermediate representation toward the activation patterns observed when generating effective feedback, thereby distilling the teacher’s guidance into the student’s latent space.
After steering, the forward propagation proceeds through the remaining layers, with subsequent feedback modulated by the steered activations, thereby altering the student behavior.

\noindent Notably, this steering procedure is training-free, model-agnostic, and operates entirely at the activation level, making it compatible with any MLLM model that provides access to its intermediate representations.

\section{Experiments}
\label{sec:experiments}

\paragraph{Baselines.}
We consider a wide range of MLLMs models, including \textsc{Qwen2.5 VL 7B}~\citep{bai2025qwen25vltechnicalreport}, \textsc{InternVL3.5 8B}~\citep{wang2025internvl3}, \textsc{Idefics3 8B}~\citep{laurencconbuilding}, and \textsc{LLaVA-OV 7B}~\citep{an2025llava}, under several configurations:
as \texttt{Teacher oracle}, models take advantage of privileged information where more memorable destination images are fed as input together with the source image, and the MLLM should only focus on generating a feedback describing their difference; as \texttt{zero-shot}, instead, models are prompted with $p_m$ to generate suggestions (see \cref{sec:method:mllm-mem}); we include \textsc{GPT-5 Mini}~\citep{OpenAI_2025_GPT5_System_Card} as a representative of proprietary models. For completeness, we compare with state-of-the-art aesthetics-specialized MLLMs trained for image perceptual evaluation, namely \textsc{Q-Instruct}~\citep{wu2024q} and \textsc{AesExpert}~\citep{huang2024aesexpert}. Finally, we also report \textsc{Editing baseline} corresponding to the empty string as feedback proposed to the editing model (see \cref{sec:method:mllm-mem}).

\paragraph{Implementation details.}
Unless stated otherwise, we use \textsc{InternVL3.5 8B}~\citep{wang2025internvl3} for both teacher and student models and employ the \benchmarkname training split to generate contrasting examples. We fix the steering layer to $l=12$ and the coefficient to $\alpha=55$, selected via tuning on a held-out subset of the training data. To ensure structured outputs, we adopt the \texttt{outlines} library~\citep{willard2023efficient} for constrained decoding (see \supmat\xspace for further details).

\subsection{Quantitative Results}
\begin{table}[t!]
    \centering
\caption{\textbf{Comparison with state-of-the-art models.} \ours performance of \inlineColorbox{methodgreen!50}{\methodname} when comparing  to several \inlineColorbox{oraclered!50}{teacher oracle}, \inlineColorbox{zsblue!19}{zero-shot} and \inlineColorbox{specializedyellow!19}{aesthetics specialized} MLLMs. \methodname achieves the best results in the considered metrics. Best results in \textbf{bold}.}
    \resizebox{0.8\linewidth}{!}{%
\begin{tabular}{l cc c}
    \toprule
    \multirow{2}{*}{\textbf{Model}} &
    \multicolumn{2}{c}{\textbf{Editing}} &
    \multirow{2}{*}{\makecell{\textbf{Perplexity}\\ ($\downarrow$)}} \\
    \cmidrule(lr){2-3}
     & IR ($\uparrow$) & RM\% ($\uparrow$) \\
     
    \midrule
        \rowcolor{gray!10}\textit{Edit model} & 0.68 & 3.72 & \textit{n.d.} \\
    \midrule
    \rowcolor{oraclered!50}\textit{Teacher oracles} &&&\\
        \modelLLaVAOV & 0.74 & 5.93 & 5.73 \\
        \modelIdefics & 0.80 & 9.84 & 29.21 \\
        \modelQwen & 0.83 & 10.16 & 2.34 \\
        \modelInternVL & 0.85 & 11.92 & 2.40 \\
    \midrule
    \rowcolor{specializedyellow!19} \textit{Aesthetics specialized} &&&\\
        \textsc{AesExpert}~\citep{huang2024aesexpert} & 0.73 & 6.67 & 5.97 \\
        \textsc{Q-Instruct}~\citep{wu2024q} & 0.73 & 5.31 & 5.36 \\
    \midrule
    \rowcolor{zsblue!19}\textit{Zero-shot baselines} &&&\\
        \modelGPTmini & 0.75 & 7.03 & \textit{n.d.} \\
        \modelLLaVAOV & 0.70 & 5.87 & 7.58 \\
        \modelIdefics & 0.73 & 6.64 & 20.19 \\
        \modelQwen & 0.68 & 4.26 & 10.23 \\
        \modelInternVL & 0.73 & 5.47 & 5.49 \\
     \midrule
     \rowcolor{methodgreen!50}\textbf{\textit{\methodname (Ours)}} & \textbf{0.80}  & \textbf{7.21} & \textbf{4.99} \\

    \bottomrule
    \end{tabular}
}
\label{tab:main-comparison}
\end{table}

Table~\ref{tab:main-comparison} reports the quantitative comparison of different MLLMs when asked for memorability feedback, as evaluated in terms of both editing metrics and perplexity. As can be noted, results highlight a consistent advantage of \methodname across both axes of evaluation. Here we only report results for \methodname when using \internvl{} model.
We observe a marked increase in IR, indicating that feedback produced by the steered model more frequently leads to edits that raise the memorability of the resulting images. This gain is further confirmed by a higher RM, showing that the relative increase in memorability is not only more frequent but also larger. \methodname yields a +5\% IR with respect to the strongest zero-shot \modelGPTmini~ and +31.81\% gain on the RM metric with respect to its base \internvl{} model. Importantly, despite its training-free nature, \methodname outperforms state-of-the-art large-scale aesthetics-specialized approaches, showcasing the benefit of the presented approach with respect to models trained on other perceptual metrics. Notably, \methodname closes the gap of training-free strategies with teacher oracle baselines that take advantage of their privileged knowledge of the scene.
Turning to the likelihood of ground-truth feedback, the lower perplexity achieved by \methodname confirms its improved alignment with human-like memorability-aware feedback: the reduced uncertainty over ground-truth feedback suggests that the steered MLLM better captures the linguistic regularities associated with memorability-increasing suggestions. Preliminary user studies in the \supmat{} confirm \methodname effectiveness and the quality of the provided feedback.

We then demonstrate that the integration of \methodname into different multimodal backbones consistently enhances their ability to generate memorability-aware feedback. Results are shown in \cref{tab:model-agnostic}. In terms of IR, \methodname yields consistent gains for all models, with the strongest improvement observed for \qwen\xspace and \llava. 
\begin{table}
    \caption{\textbf{Generalization to different MLLMs.} \ours performance of \inlineColorbox{methodgreen!50}\methodname when applied to different architectures. \methodname generalizes to different models, enhancing their ability to produce memorability feedback. 
    }
    \centering
    \resizebox{0.95\linewidth}{!}{%
\begin{tabular}{l cc c}
    \toprule
    \multirow{2}{*}{\textbf{Model}} &
    \multicolumn{2}{c}{\textbf{Editing}} &
    \multirow{2}{*}{\makecell{\textbf{Perplexity}\\ ($\downarrow$)}} \\
    \cmidrule(lr){2-3}
     & IR ($\uparrow$) & RM\% ($\uparrow$)\\
    
    \midrule
        \modelLLaVAOV & 0.70 & \textbf{5.87} & \textbf{7.58} \\
        \rowcolor{methodgreen!50} \textit{\textbf{\methodname-\textsc{LLaVA}}} & \textbf{0.73}~\gain{4.29}  & 5.04~\drop{14.14} & 14.05~\perplexitydrop{85.36}  \\
        \midrule
        \modelIdefics & 0.73 & 6.64 & 20.19 \\
        \rowcolor{methodgreen!50} \textit{\textbf{\methodname-\textsc{Idefics}}}& \textbf{0.75}~\gain{2.74}  & \textbf{6.69}~\gain{0.75} & \textbf{19.81}~\perplexitygain{1.88} \\
        \midrule
        \modelQwen & 0.68 & 4.26 & \textbf{10.23} \\
        \rowcolor{methodgreen!50}\textit{\textbf{\methodname-\textsc{Qwen}}} & \textbf{0.74}~\gain{8.82}  & \textbf{5.49}~\gain{28.87}  & 13.90~\perplexitydrop{35.87} \\
        \midrule
        \modelInternVL & 0.73 & 5.47 & 5.49 \\
        \rowcolor{methodgreen!50}\textit{\textbf{\methodname-\textsc{InternVL}}} & \textbf{0.80}~\gain{9.59}  & \textbf{ 7.21}~\gain{31.81}  & \textbf{4.99}~\perplexitygain{9.11} \\

    \bottomrule
    \end{tabular}
}
\label{tab:model-agnostic}
\end{table}

\subsection{Qualitative Evaluation}
\begin{figure*}[t!h!]
    \centering
    \includegraphics[width=.9\linewidth]{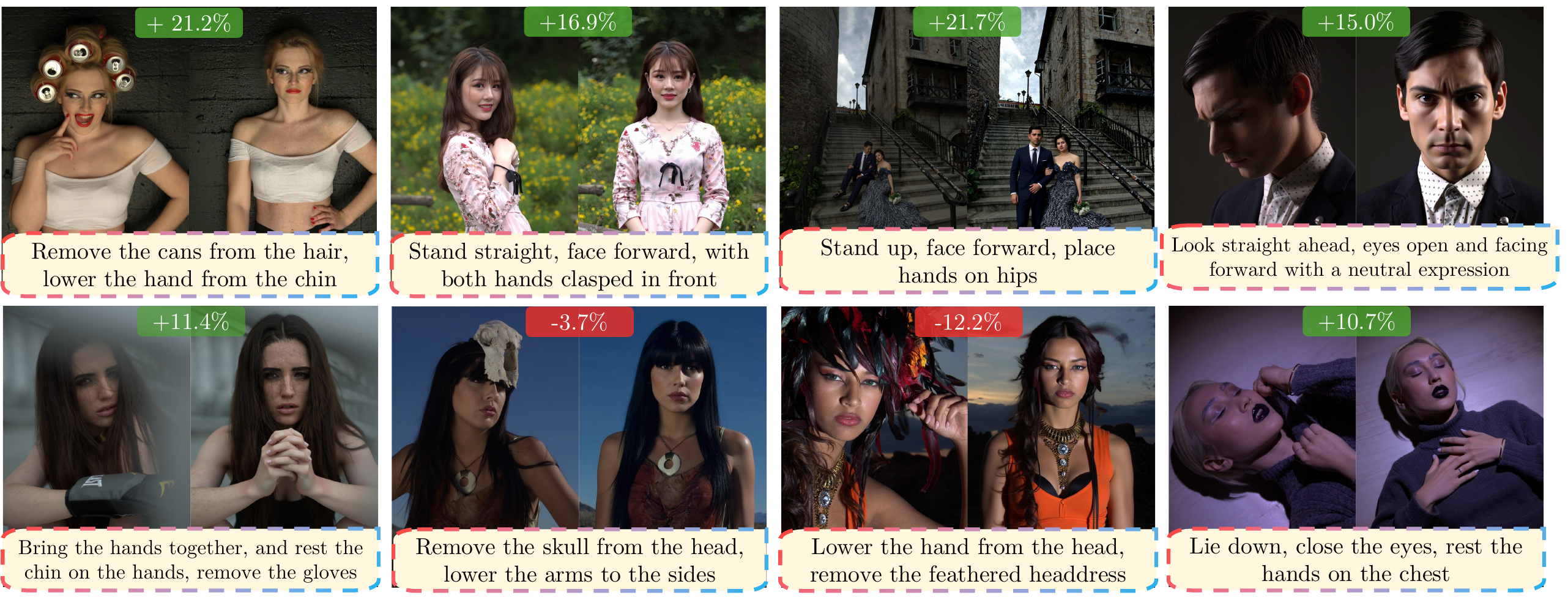}
    \caption{\textbf{Qualitative feedback from \methodname.} For each source image (left), the model provides natural-language feedback (bottom) that is applied to produce the destination image (right). Each score represents the Relative Memorability (RM), indicating how suggested feedback affects memorability. \methodname provides human-interpretable and actionable feedback that translates into semantic changes for overall improved memorability. Observed failure cases propose to remove out-of-context elements.}
    \label{fig:feedback-qualitatives}
\end{figure*}

We qualitatively analyze the feedback provided by \methodname in \cref{fig:feedback-qualitatives}, where source images observed by the model (left) are shown with the provided natural-language suggestions (bottom) and the imagined destination image (right), as generated by the in-context editing model. The examples highlight the variety of suggestions the model proposes, ranging from fine-grained compositional adjustments, such as altering gaze direction, pose, or hand position, to semantic interventions involving object removal or face expression change. Feedback is naturally interpretable and actionable, expressed in concise textual instructions (mostly involving verbs \textit{``Bring'', ``Stand'', ``Remove''}) that can be directly implemented, effectively verbalizing how to take a memorable picture. Interestingly, cases in the figure also expose trade-offs between normalization and distinctiveness. In line with previous memorability studies~\citep{goetschalckx2019ganalyze}, positive cases often relate to conventional photographic strategies (\textit{e.g.}, centered framing, and minimal occlusion). Conversely, failure cases show the negative effect of removing semantically out-of-context elements (\textit{e.g.}, skulls, feathered headdresses), underscoring the dual nature of memorable images, where both clarity and the extrinsic notion distinctiveness~\citep{bylinskii2015intrinsic} shape the \ours task.

\begin{figure}
    \centering
\includegraphics[width=.9\linewidth]{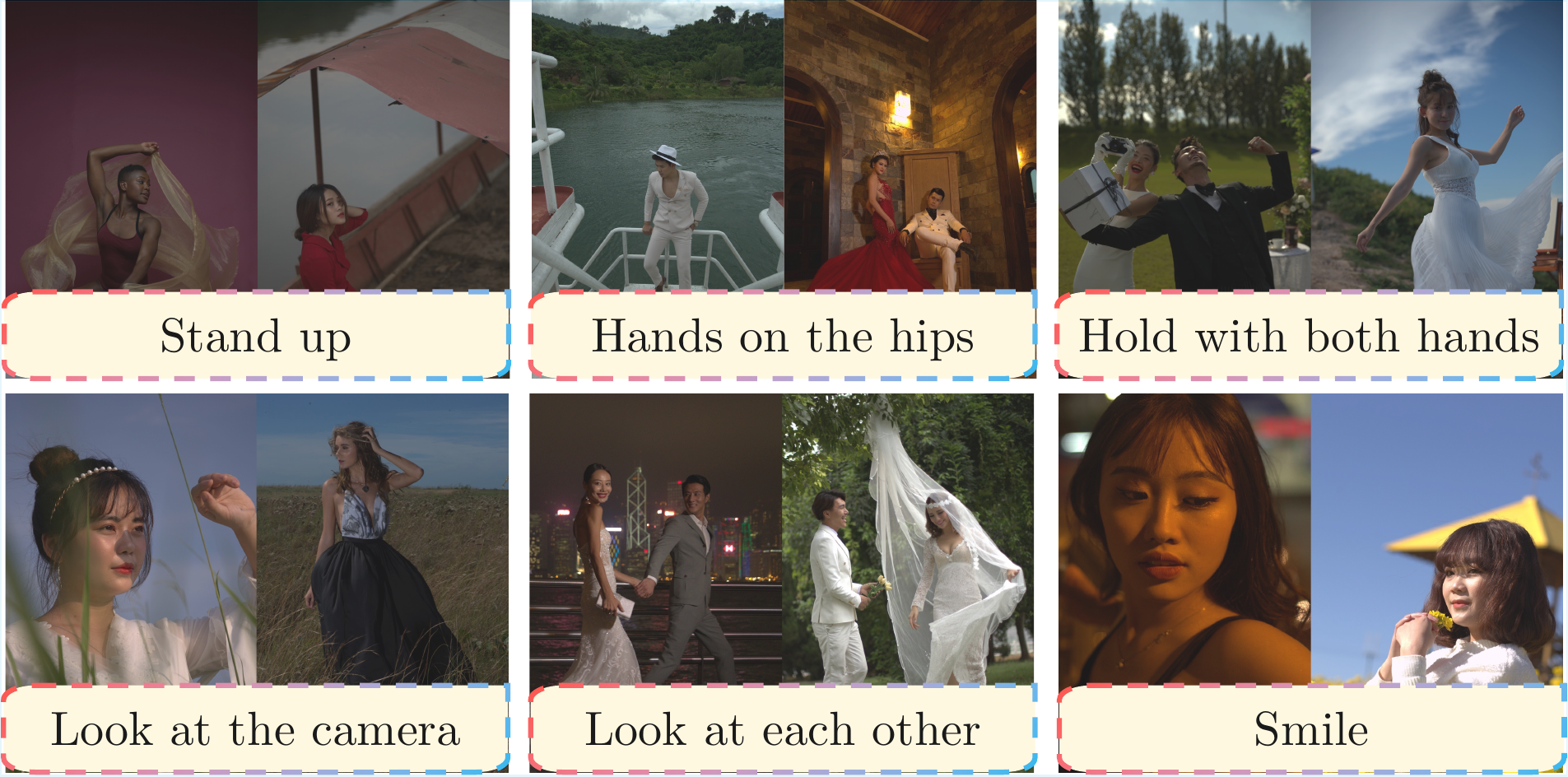}
    \caption{\textbf{Common feedback patterns on source images.} \methodname favors symmetric and socially connected compositions, reflecting principles of human photography.}
    \label{fig:common-feedback}
\end{figure}

\paragraph{Common feedback patterns.} In \cref{fig:common-feedback}, we analyze the most recurrent feedback patterns that \methodname associates with improved memorability. Interestingly, these suggestions reveal an emergent understanding of photographic composition and social engagement. Many instructions promote symmetry and balance, such as \textit{``hold with both hands''} or `\textit{`hands on the hips''}, which encourage centered and symmetric poses that naturally guide the viewer’s attention toward the subject~\citep{bylinskii2015intrinsic, kumar2023eye}. Others focus on directing the subjects' gaze, such as \textit{``look at the camera''} or \textit{``look at each other''}, reinforcing its role as emotionally resonant cue.

\subsection{Ablation Study}
\begin{figure}
    \centering
\includegraphics[width=0.75\linewidth]{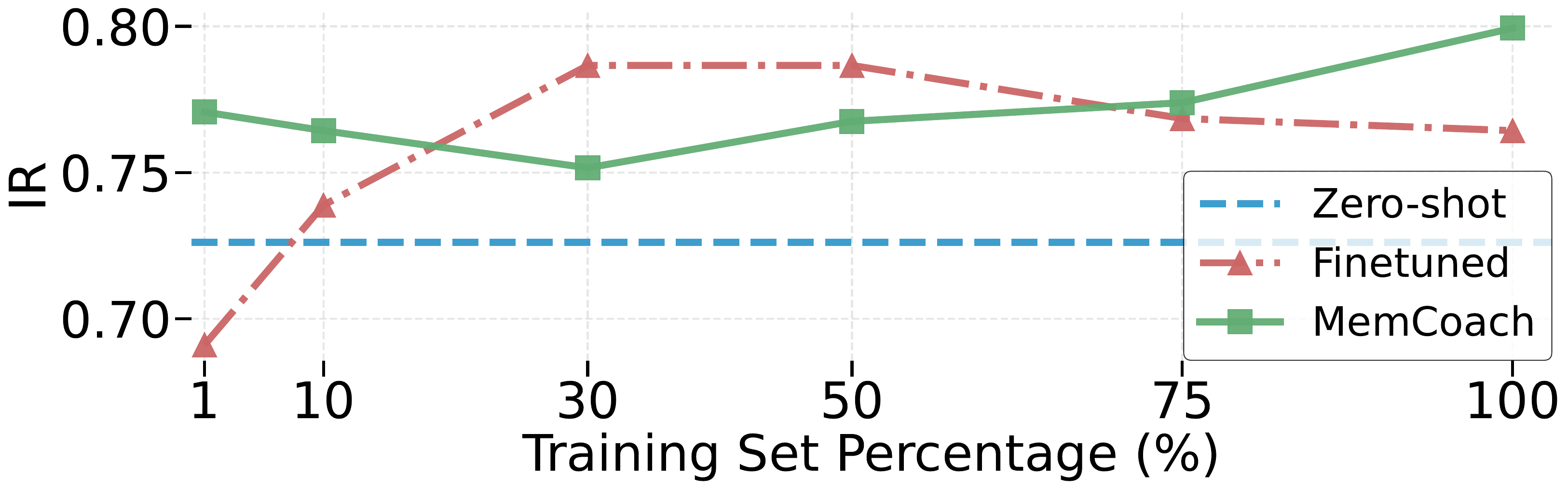}
\includegraphics[width=0.75\linewidth]{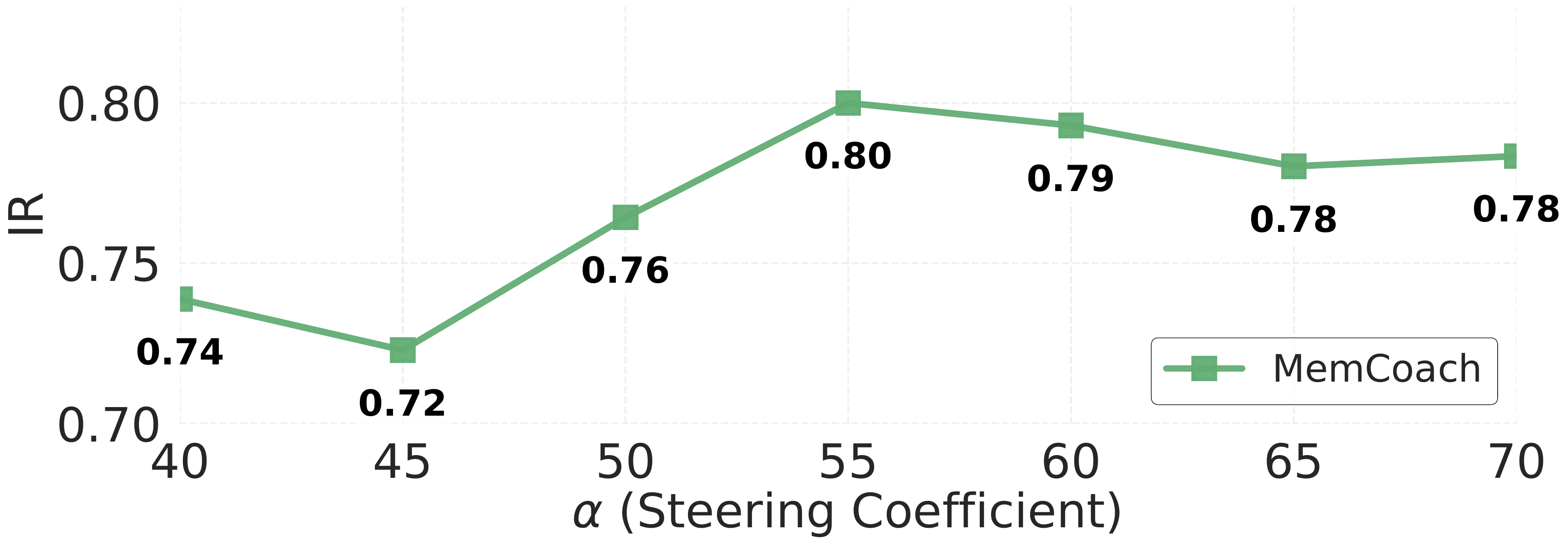}
    \caption{\textbf{Data efficiency.} \textbf{Top:} performance \textit{vs} number of training/contrasting samples. 
    \textbf{Bottom:} performance 
    \textit{vs} coefficient $\alpha$.}
    \label{fig:data-efficiency}
\end{figure}

 \paragraph{Data efficiency of steering.}
\Cref{fig:data-efficiency}-top compares the improvement ratio as a function of the available training data. In the low-data regime, \methodname{} consistently outperforms Low-Rank~\citep{hu2022lora} fine-tuning, showing that steering requires far fewer samples to capture memorability-relevant directions. 
With only $1\%$ of the training data, \methodname{} already reaches performance on par with full-data fine-tuning, while maintaining stable gains as more data become available.

\paragraph{Impact of main components.} \cref{tab:ablation} analyzes the main design choices underlying \methodname. \textsc{Qwen-Contrasting} reports model performance when the memorability-aware feedback in the contrasting data generation is extracted from a different teacher (\qwen). As can be noted, steering continues to provide a positive effect, though with reduced marginal benefit compared to the \internvl{}. Confirming the importance of per-sample contrast, the \textsc{Diff(Mean)} variant, which averages activations before differencing, yields lower editing performance ($6.64$ RM) than our subtraction-before-averaging formulation ($7.21$ RM), presented in Eq. \ref{eq:steering-compute}. Finally, ~\cref{fig:data-efficiency}-bottom ablates the steering parameter $\alpha$ in terms of IR: performance improvement is initially observed with increasing coefficient values, with performance saturating with larger alphas.

\begin{table}[ht]
\centering
\caption{\textbf{Ablation analysis.} \ours performance of \inlineColorbox{methodgreen!50}\methodname when ablating on the contrasting data teacher and steering vector computation. 
}
\label{tab:ablation}
\resizebox{0.65\linewidth}{!}{%
\begin{tabular}{lccc}
\toprule
    \multirow{2}{*}{\textbf{Model}} &
    \multicolumn{2}{c}{\textbf{Editing}} &
    \multirow{2}{*}{\makecell{\textbf{Perplexity}\\ ($\downarrow$)}} \\
    \cmidrule(lr){2-3}
     & IR ($\uparrow$) &  RM\% ($\uparrow$) \\
\midrule
\textsc{Qwen-contrasting} & 0.73 & 5.68 & 5.13\\
\textsc{Diff(Mean)} & 0.78 & 6.64 & \textbf{4.39}\\
\midrule

\rowcolor{methodgreen!50}\textit{\textbf{\methodname (Ours)}} & \textbf{0.80} &\textbf{7.21} & 4.99\\
\bottomrule
\end{tabular}
}
\end{table}

\section{Conclusion}
We introduced the challenging problem of \ourslong, a new task that shifts the study of memorability from passive prediction to actionable guidance. To foster future research on the setting, we present \benchmarkname along with \ours evaluation metrics to assess the quality of provided feedback. We proposed \methodname, a novel model-agnostic activation steering framework that distills how to improve the memorability of an image from an oracle teacher model to a student MLLM, aiming to provide natural-language feedback at capture time. Experimental validation of the approach demonstrates that steering multimodal large language models towards memorability-aware activations yields more effective and human-aligned feedback than zero-shot strategies, while requiring only minimal data. Beyond memorability, our findings suggest that activation steering offers a general and efficient route to endow MLLMs with perceptual skills, paving the way for future research on interactive and explainable visual guidance systems.

\clearpage
\section*{Acknowledgments}

We acknowledge ISCRA for awarding this project access to the LEONARDO supercomputer, owned by the EuroHPC Joint Undertaking, hosted by CINECA (Italy). This work was supported by the Ministero delle Imprese e del Made in Italy (IPCEI Cloud DM 27 giugno 2022 – IPCEI-CL-0000007) and European Union (Next Generation EU), the EU Horizon ELIAS (No. 101120237), and ELLIOT (No. 101214398). This work was carried out in the Vision and Learning joint laboratory of FBK and UniTN. Francesco Laiti is supported by PNRR funding (Innovative Doctorates program).

{
    \small
    \bibliographystyle{ieeenat_fullname}
    \bibliography{main}
}

\appendix
\clearpage
\maketitlesupplementary

In this supplementary material, we provide additional details on \benchmarkname and \methodname. Section \ref{sec:membench_details} presents qualitative examples from \benchmarkname dataset and describes its construction pipeline. Section \ref{suppl:membench_tech_details} gives additional implementation details of our proposed MemCoach method with a discussion regarding potential implications of our work, while Section \ref{sec:human-studies-appendix} provides preliminary user study experiments. Finally, in Section \ref{sec:add-analysis-appendix}, we demonstrate the consistency of our framework across different editing models and memorability predictors, including an analysis of feedback quality generated by MemCoach.

\section{\benchmarkname Additional Details} \label{sec:membench_details}

\subsection{Data Examples}\label{sec:membench_qualitatives}
\label{sec:supp_qualitative_res_membench}
\begin{figure*}[]
\centering
\begin{minipage}{0.16\textwidth}
    \centering
    \begin{tikzpicture}
        \node[draw=red, line width=1pt] (imgA) {\includegraphics[width=\linewidth]{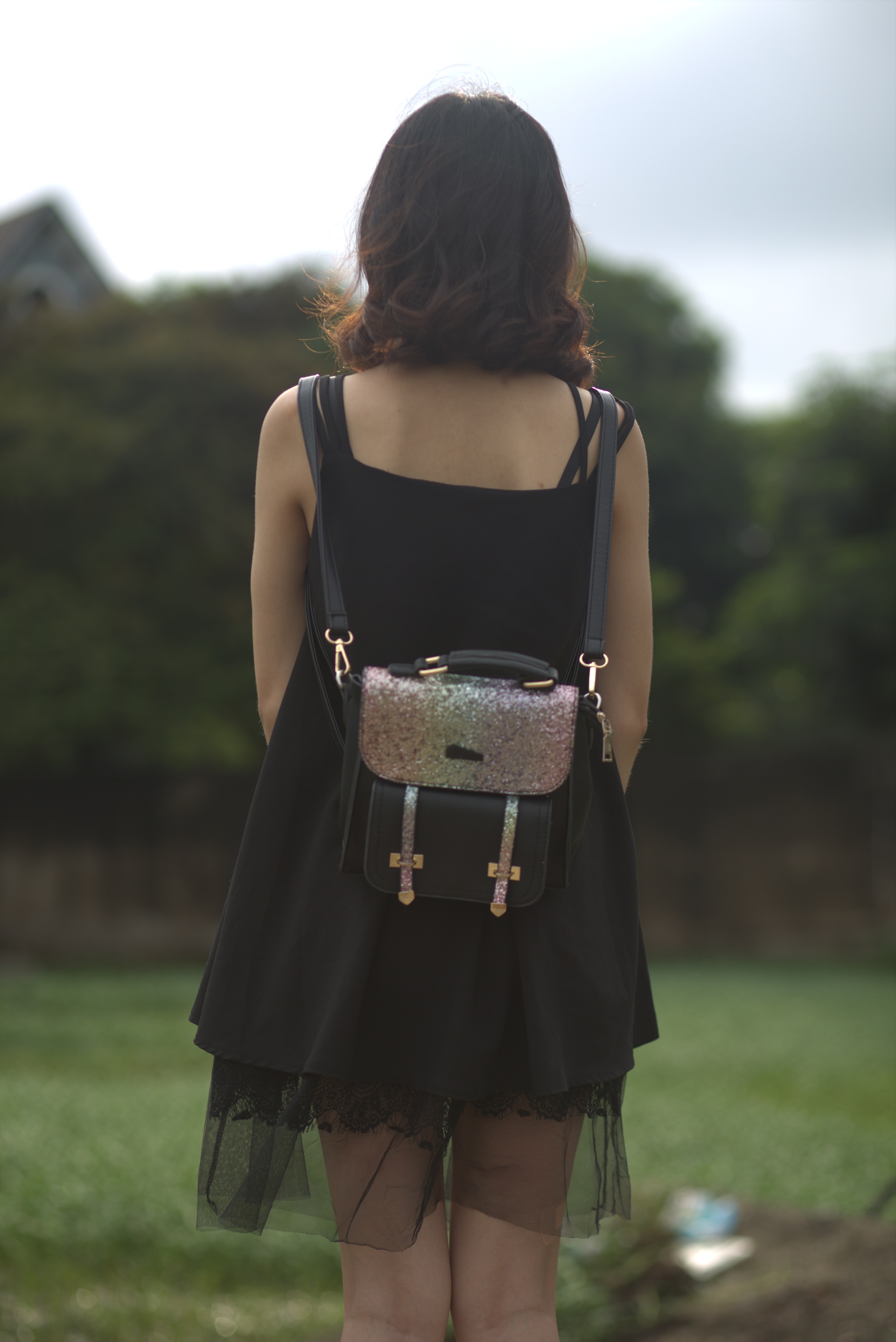}};
        \node[below=0.5mm of imgA] {Mem score: 0.525};
    \end{tikzpicture}
\end{minipage}%
\hfill
\begin{minipage}{0.16\textwidth}
    \centering
    \begin{tikzpicture}
        \node[draw=blue, line width=1pt] (imgB) {\includegraphics[width=\linewidth]{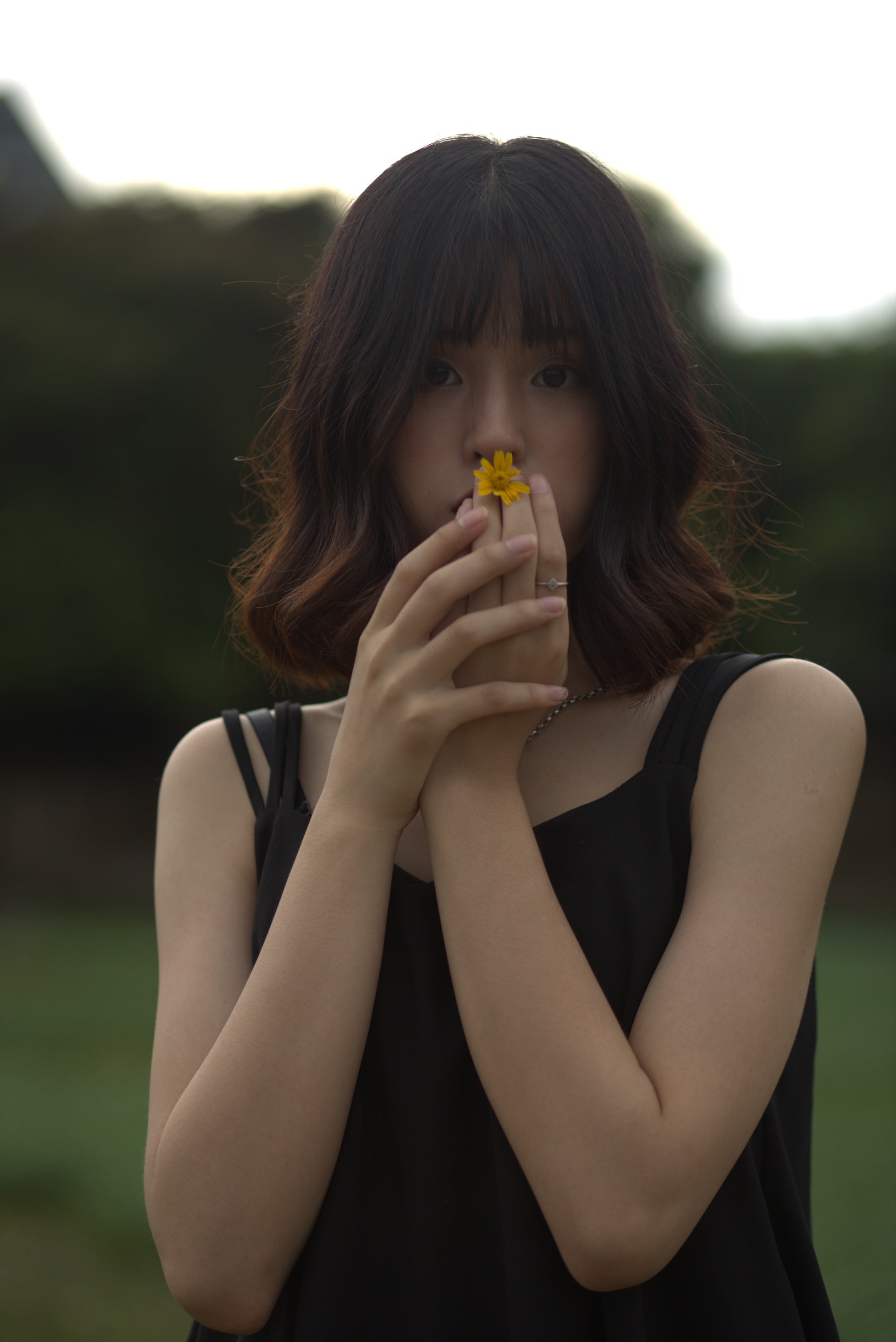}};
        \node[below=0.5mm of imgB] {Mem score: 0.992};
    \end{tikzpicture}
\end{minipage}
\hfill
\begin{minipage}{0.45\textwidth}
    \textbf{Feedback $a$:}
    \begin{enumerate}
        \item Rotate the perspective to face forward.
        \item Bring the hands up to cover the mouth.
        \item Hold a small yellow flower between the fingers.
        \item Adjust the hair to frame the face evenly.
    \end{enumerate}
\end{minipage}
\hrule
\vspace{1em}

\centering
\begin{minipage}{0.16\textwidth}
    \centering
    \begin{tikzpicture}
        \node[draw=red, line width=1pt] (imgA) {\includegraphics[width=\linewidth]{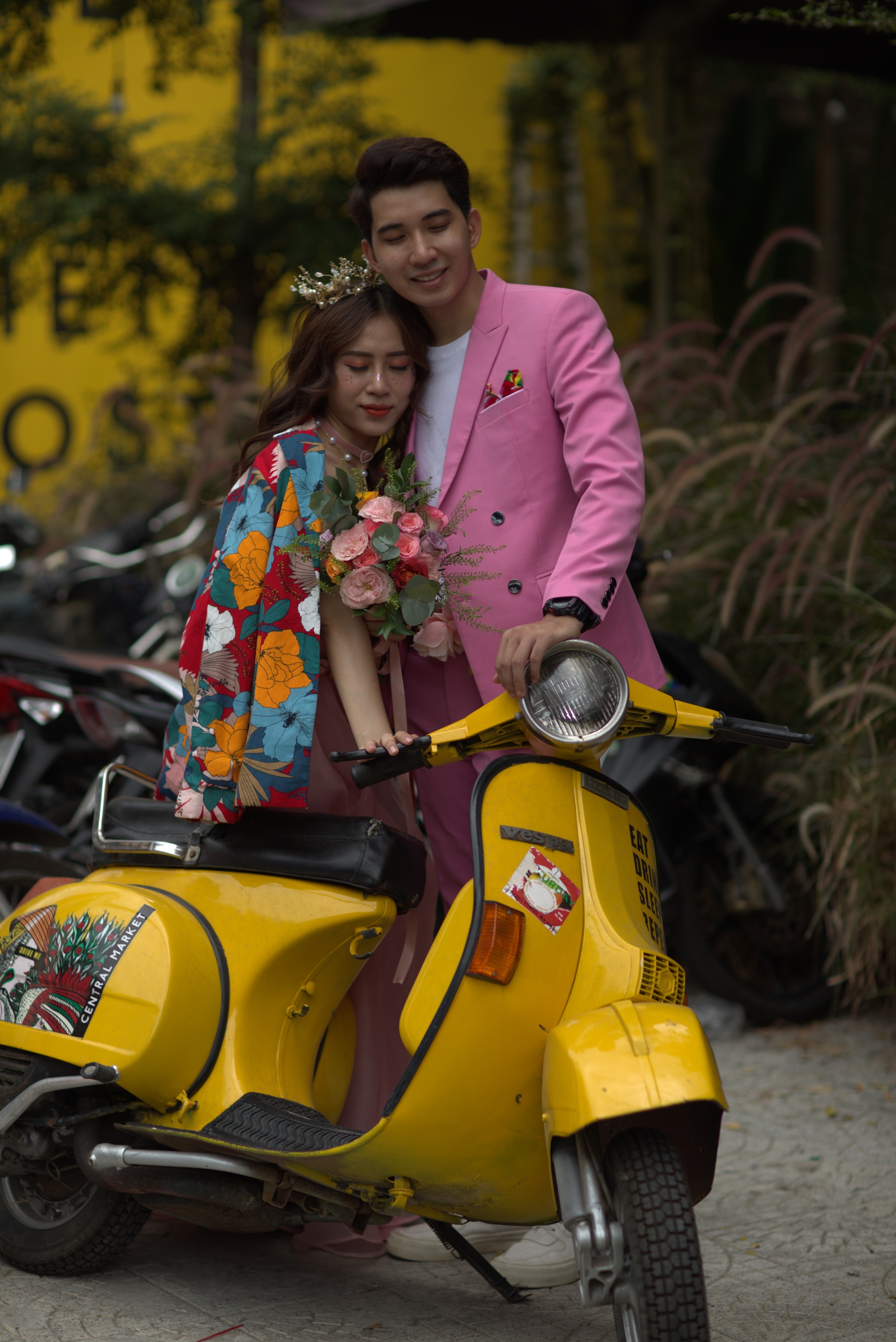}};
        \node[below=0.5mm of imgA] {Mem score: 0.753};
    \end{tikzpicture}
\end{minipage}%
\hfill
\begin{minipage}{0.16\textwidth}
    \centering
    \begin{tikzpicture}
        \node[draw=blue, line width=1pt] (imgB) {\includegraphics[width=\linewidth]{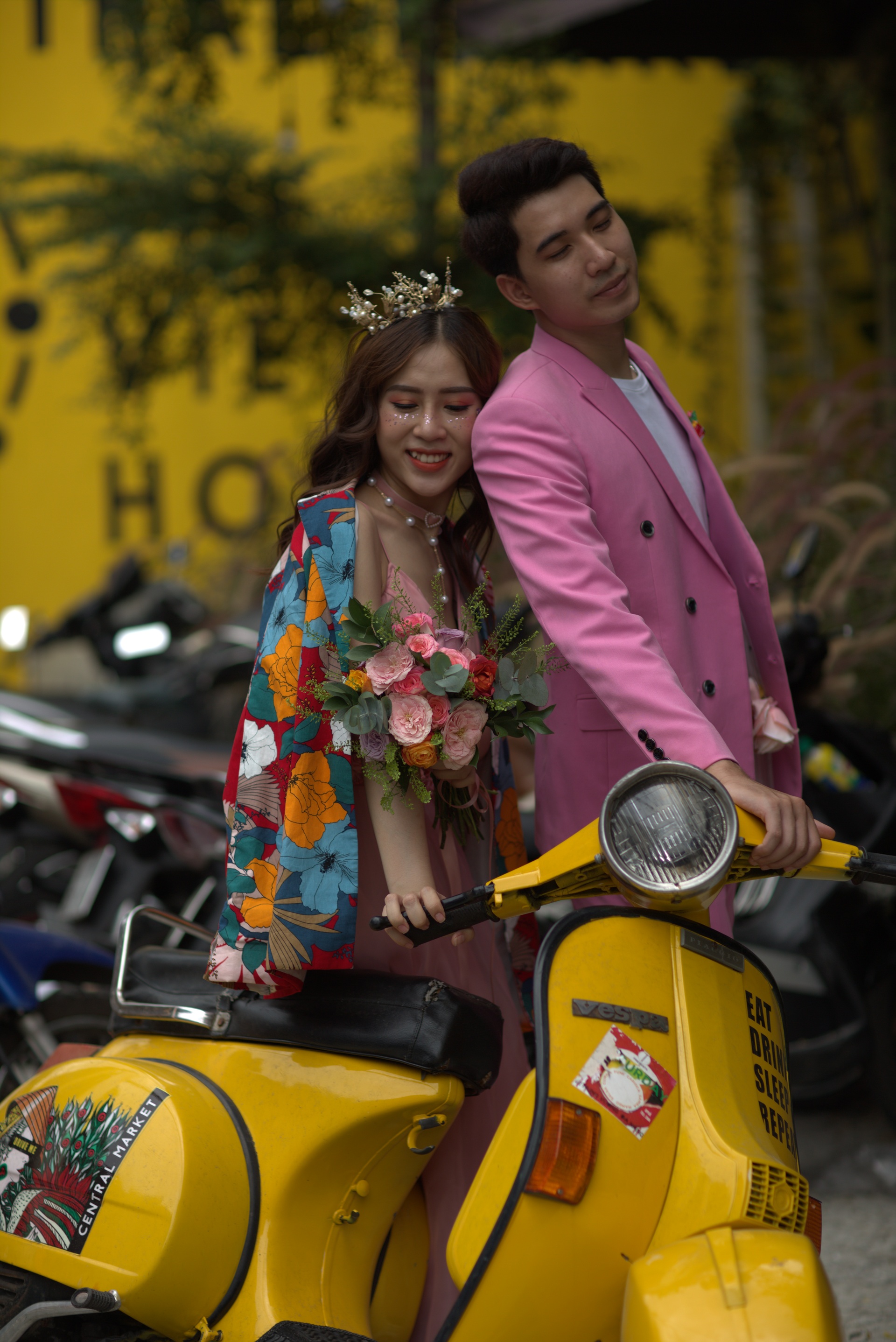}};
        \node[below=0.5mm of imgB] {Mem score: 0.803};
    \end{tikzpicture}
\end{minipage}
\hfill
\begin{minipage}{0.45\textwidth}
    \textbf{Feedback $a$:}
    \begin{enumerate}
    \item Adjust the position of the person on the left to face forward with a slight smile.
    \item Raise the head of the person on the right and have them look slightly to the side with a gentle smile.
    \item Ensure both individuals are standing upright and close together, with the person on the right holding the handlebars of the scooter.
    \item Maintain the floral arrangement and attire of both individuals as they are.
    \end{enumerate}
\end{minipage}

\hrule
\vspace{1em}

\centering
\begin{minipage}{0.16\textwidth}
    \centering
    \begin{tikzpicture}
        \node[draw=red, line width=1pt] (imgA) {\includegraphics[width=\linewidth]{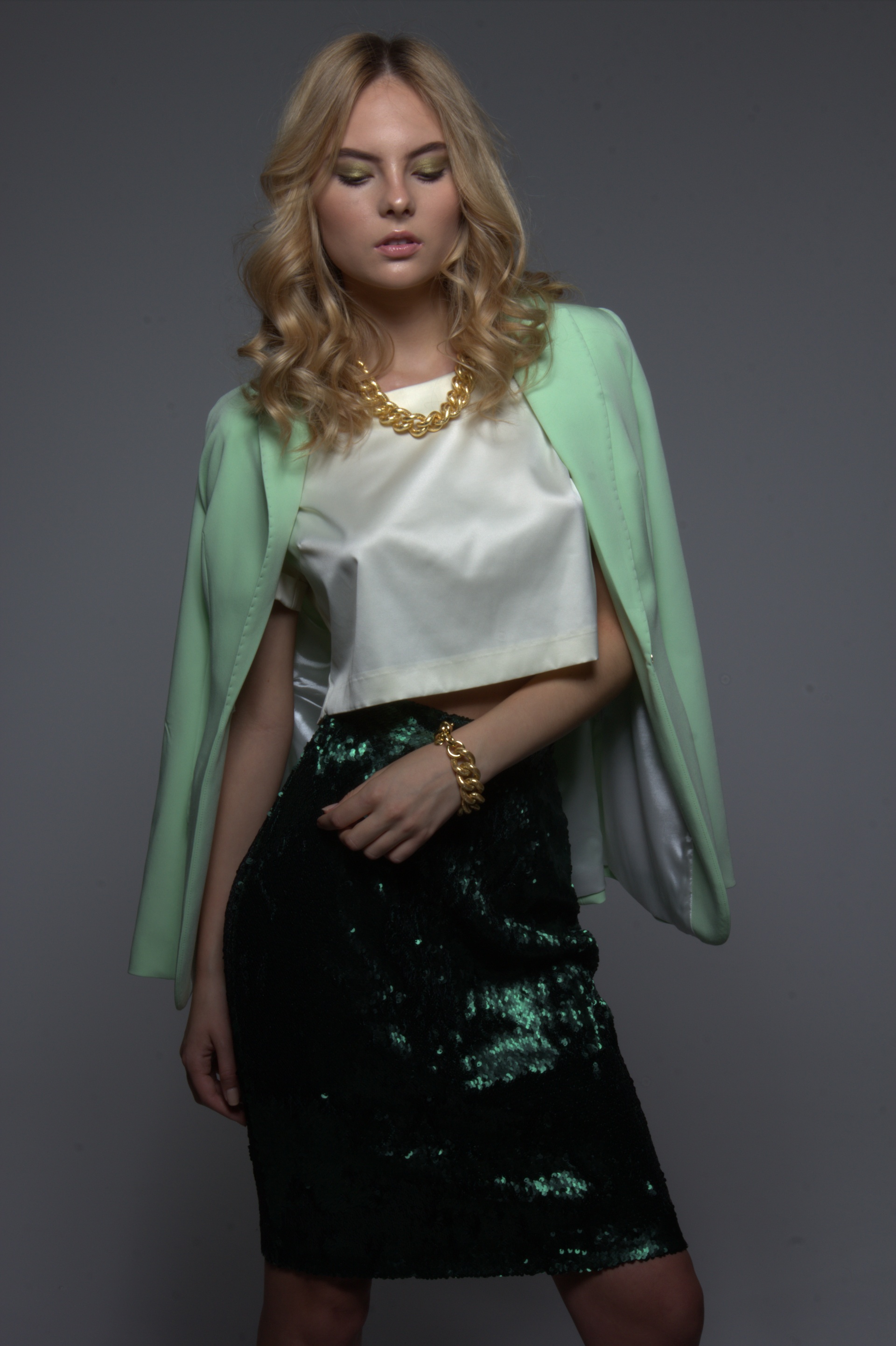}};
        \node[below=0.5mm of imgA] {Mem score: 0.885};
    \end{tikzpicture}
\end{minipage}%
\hfill
\begin{minipage}{0.16\textwidth}
    \centering
    \begin{tikzpicture}
        \node[draw=blue, line width=1pt] (imgB) {\includegraphics[width=\linewidth]{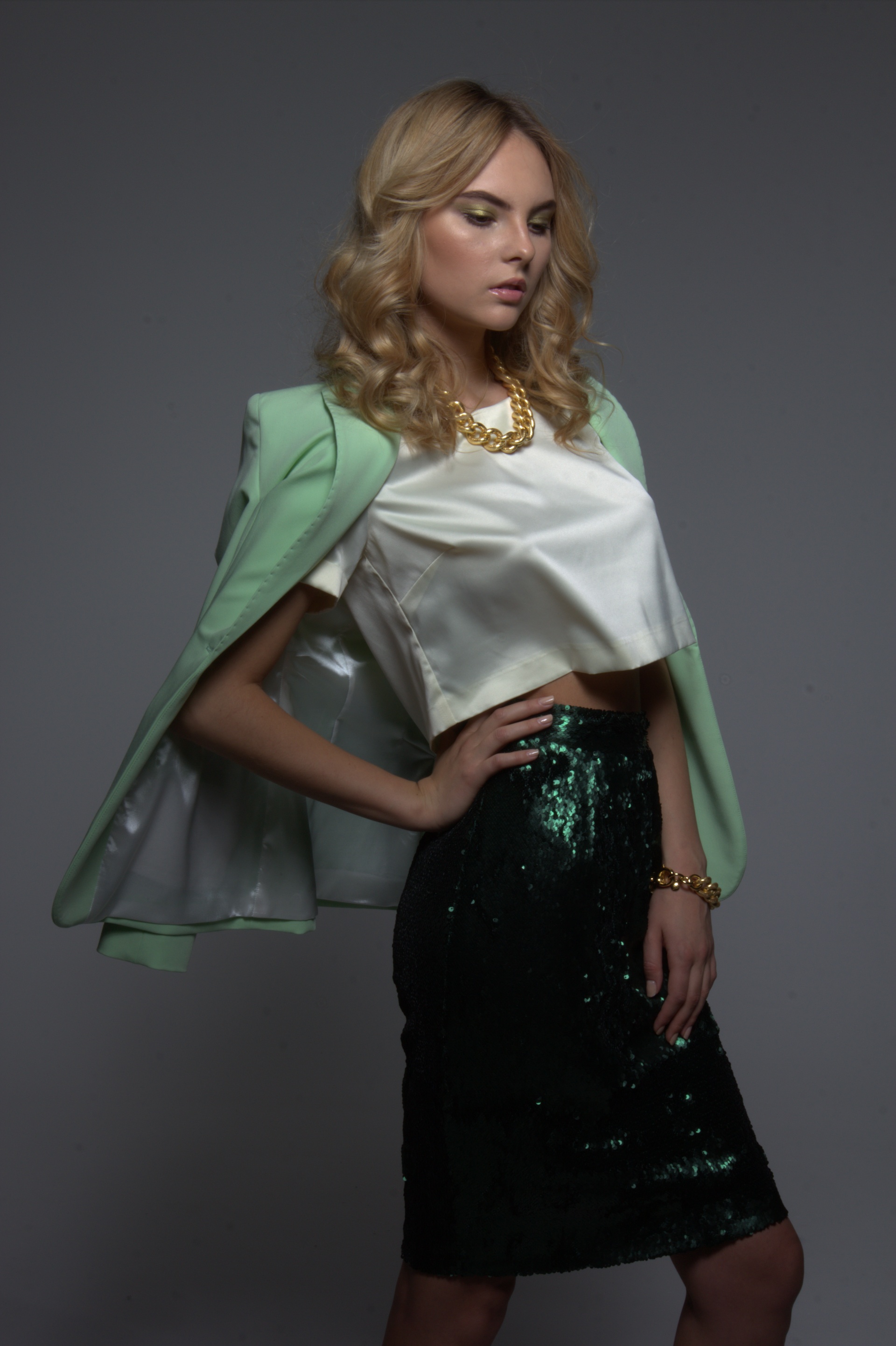}};
        \node[below=0.5mm of imgB] {Mem score: 0.965};
    \end{tikzpicture}
\end{minipage}
\hfill
\begin{minipage}{0.45\textwidth}
    \textbf{Feedback $a$:}
    \begin{enumerate}
    \item Adjust the position of the left arm to rest on the hip, ensuring the hand is placed firmly on the hip bone.
    \item Shift the weight slightly to the left leg to create a more dynamic stance.
    \item Tilt the head slightly to the right to add a subtle angle to the gaze.
    \item Ensure the right hand is relaxed by the side, with fingers gently curved.
    \item Adjust the lighting to maintain consistency in shadows and highlights across the outfit.
    \end{enumerate}
\end{minipage}

\hrule
\vspace{1em}

\centering
\begin{minipage}{0.25\textwidth}
    \centering
    \begin{tikzpicture}
        \node[draw=red, line width=1pt] (imgA) {\includegraphics[width=\linewidth]{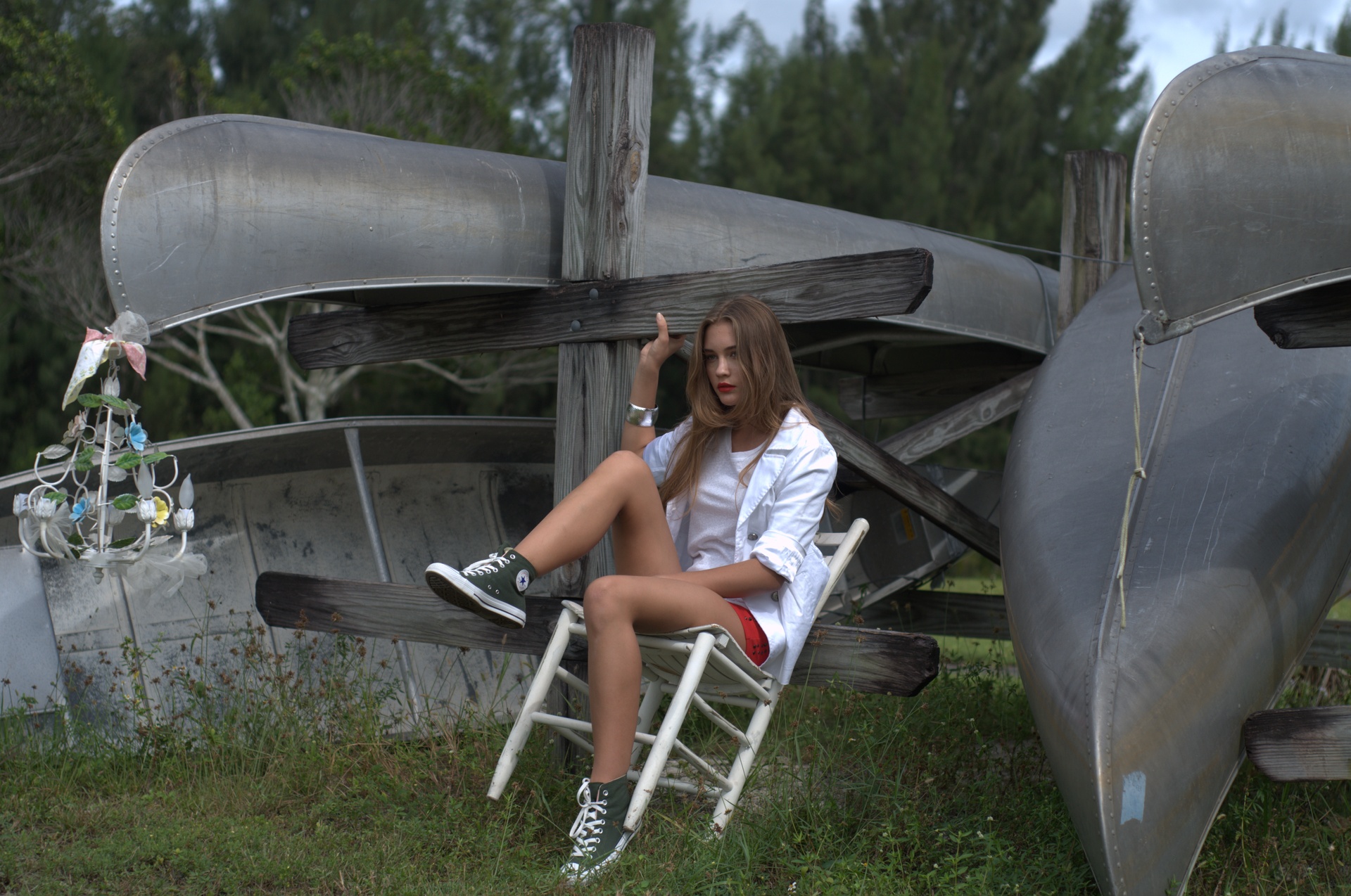}};
        \node[below=0.5mm of imgA] {Mem score: 0.578};
    \end{tikzpicture}
\end{minipage}%
\hfill
\begin{minipage}{0.16\textwidth}
    \centering
    \begin{tikzpicture}
        \node[draw=blue, line width=1pt] (imgB) {\includegraphics[width=\linewidth]{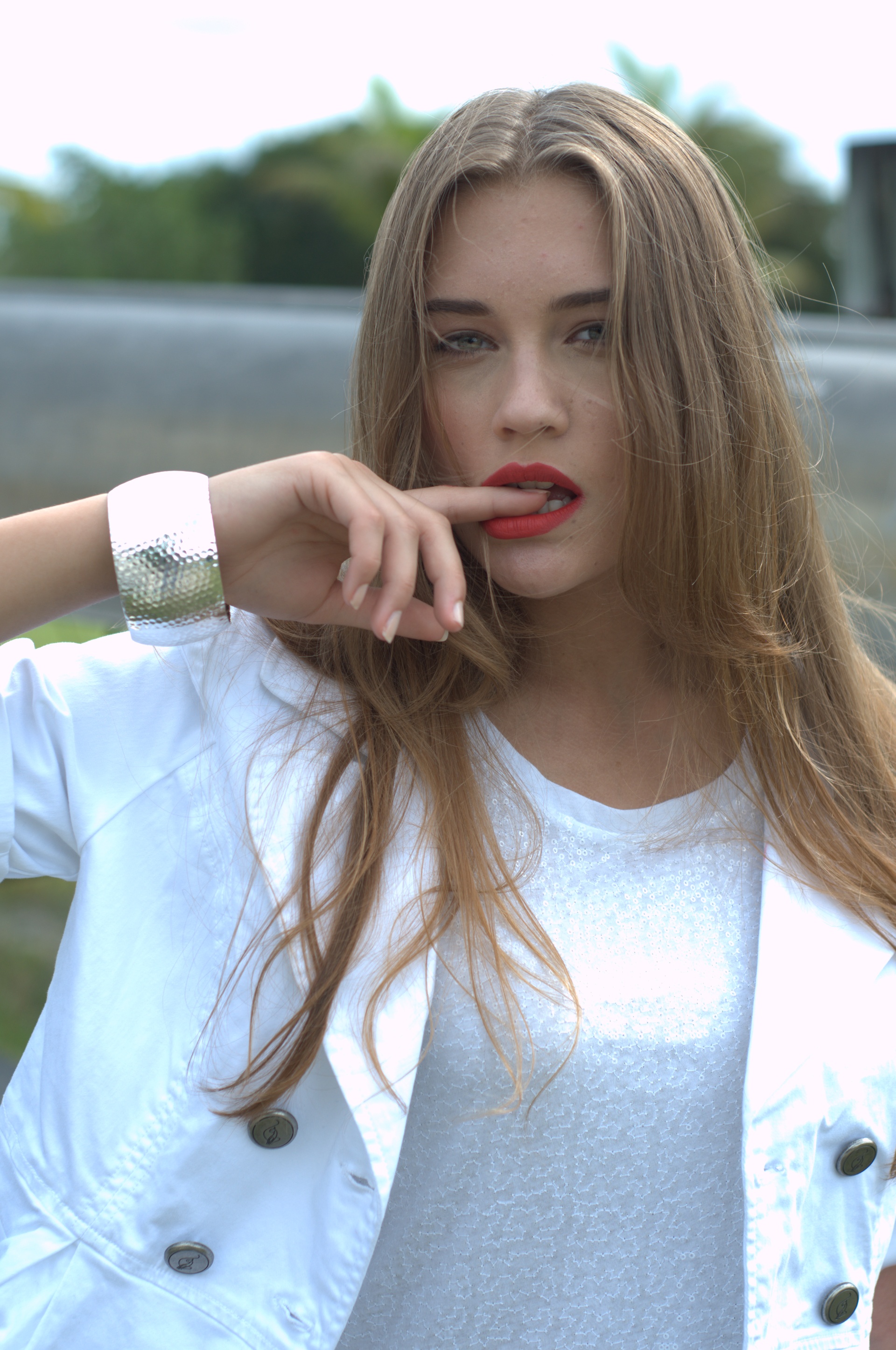}};
        \node[below=0.5mm of imgB] {Mem score: 0.981};
    \end{tikzpicture}
\end{minipage}
\hfill
\begin{minipage}{0.45\textwidth}
    \textbf{Feedback $a$:}
    \begin{enumerate}
    \item Crop the image to focus on the upper body of the person, removing the background elements such as the airplane and the chair.
    \item Adjust the angle to a closer, more intimate shot, centering the person's face and upper torso.
    \item Ensure the person's hand is positioned near the face, with the index finger touching the lips, maintaining the same pose but in a tighter frame.
    \item Remove any elements that are not part of the person's attire or immediate surroundings, such as the grass and distant trees.
   \item Enhance the lighting to create a softer, more even illumination on the person's face and clothing.
    \end{enumerate}
\end{minipage}
\caption{A set of qualitative examples from MemBench.}
\label{fig:suppl_qualitatives}
\end{figure*}

\begin{figure*}[]
\centering
\begin{minipage}{0.13\textwidth}
    \centering
    \begin{tikzpicture}
        \node[draw=red, line width=1pt] (imgA) {\includegraphics[width=\linewidth]{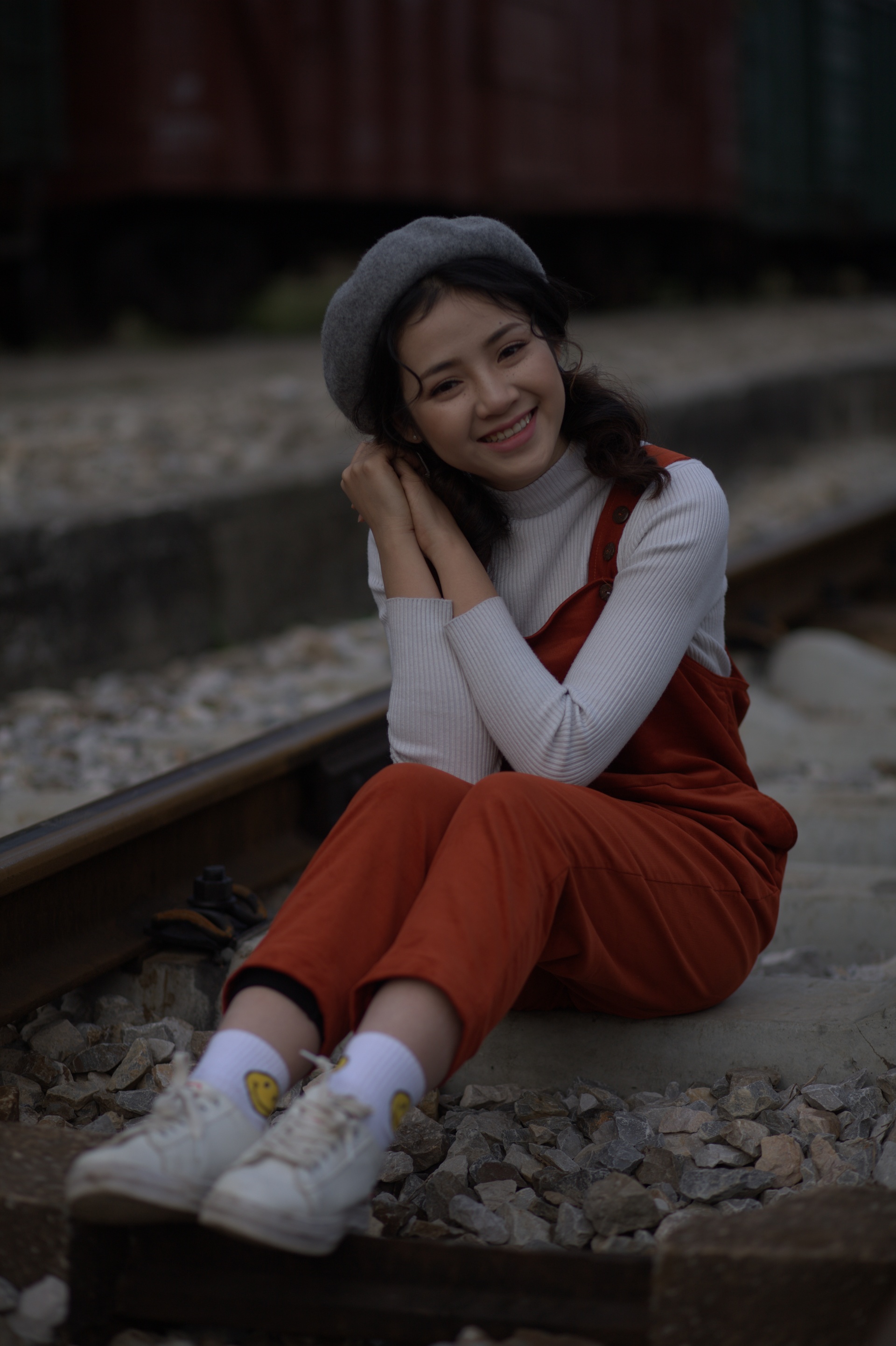}};
        \node[below=0.5mm of imgA] {Mem score: 0.716};
    \end{tikzpicture}
\end{minipage}%
\hfill
\begin{minipage}{0.25\textwidth}
    \centering
    \begin{tikzpicture}
        \node[draw=blue, line width=1pt] (imgB) {\includegraphics[width=\linewidth]{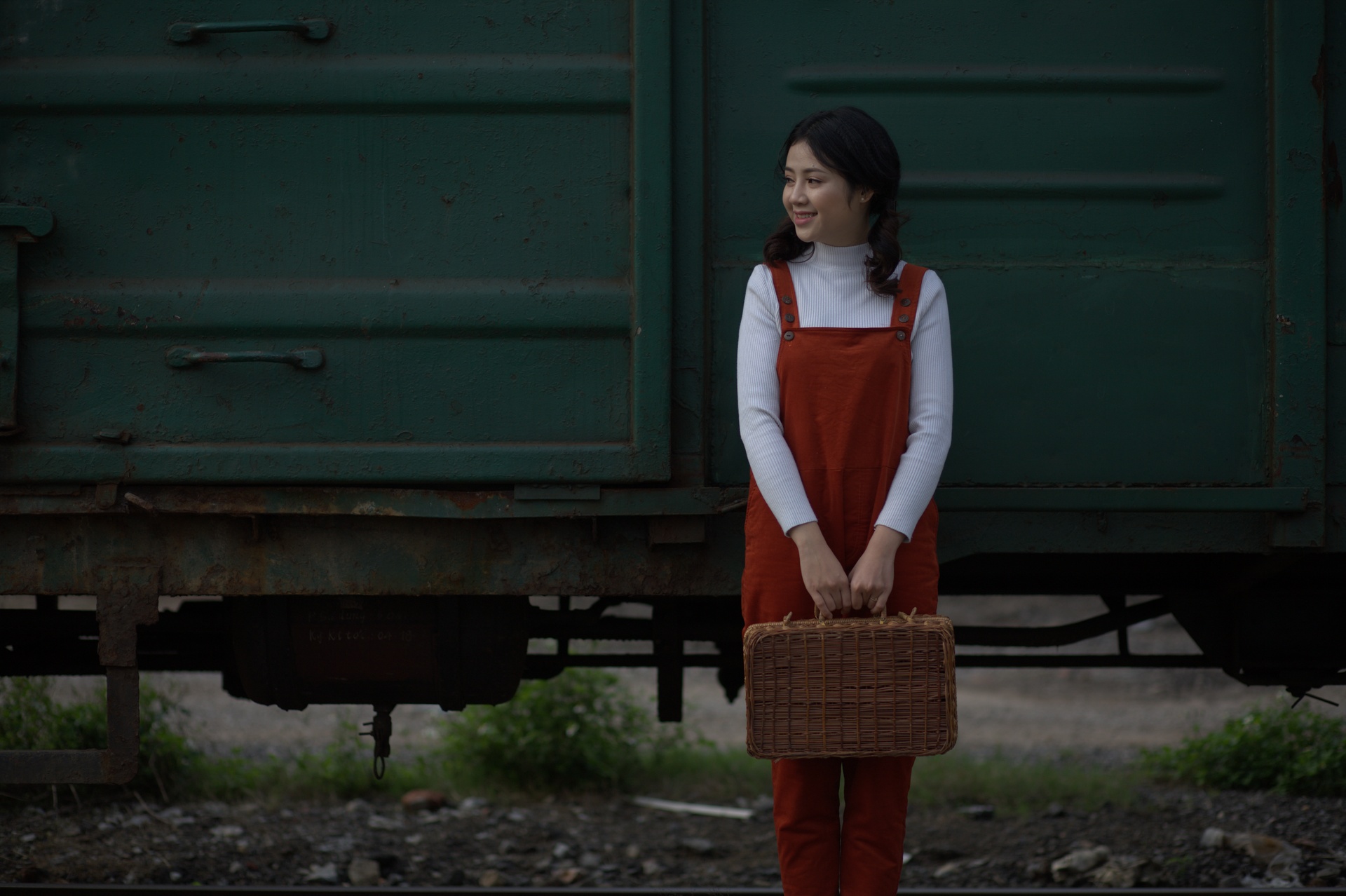}};
        \node[below=0.5mm of imgB] {Mem score: 0.980};
    \end{tikzpicture}
\end{minipage}
\hfill
\begin{minipage}{0.45\textwidth}
    \textbf{Feedback $a$:}
    \begin{enumerate}
        \item Stand up from the seated position on the railway track.
        \item Hold a wicker basket with both hands in front of the body.
        \item Adjust the hair to fall naturally over the shoulders.
        \item Shift the gaze to the side while maintaining a smile.
        \item Ensure the background changes to a green train car with visible rust and metal details.
    \end{enumerate}
\end{minipage}
\hrule
\vspace{1em}

\centering
\begin{minipage}{0.25\textwidth}
    \centering
    \begin{tikzpicture}
        \node[draw=red, line width=1pt] (imgA) {\includegraphics[width=\linewidth]{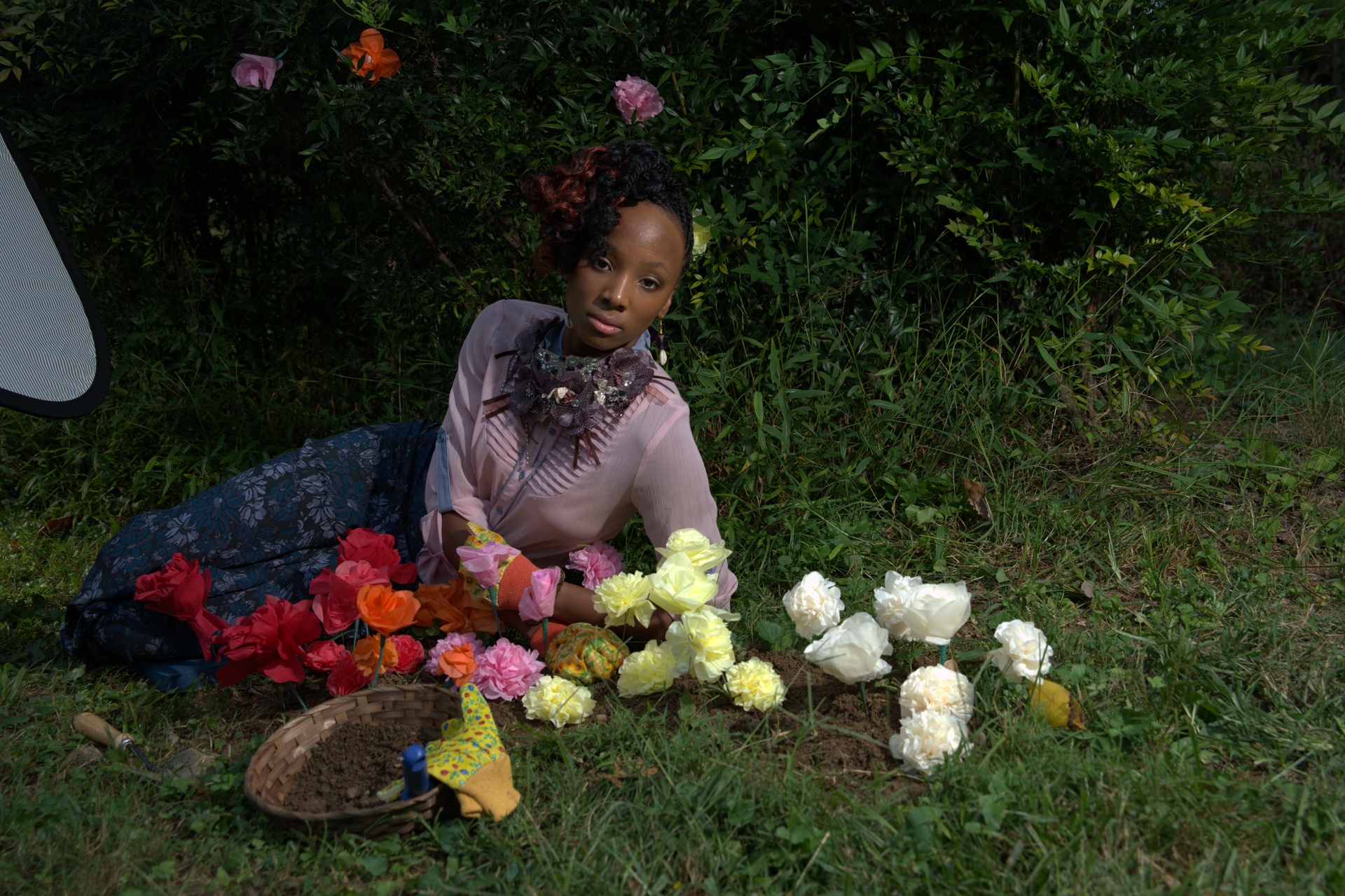}};
        \node[below=0.5mm of imgA] {Mem score: 0.710};
    \end{tikzpicture}
\end{minipage}%
\hfill
\begin{minipage}{0.25\textwidth}
    \centering
    \begin{tikzpicture}
        \node[draw=blue, line width=1pt] (imgB) {\includegraphics[width=\linewidth]{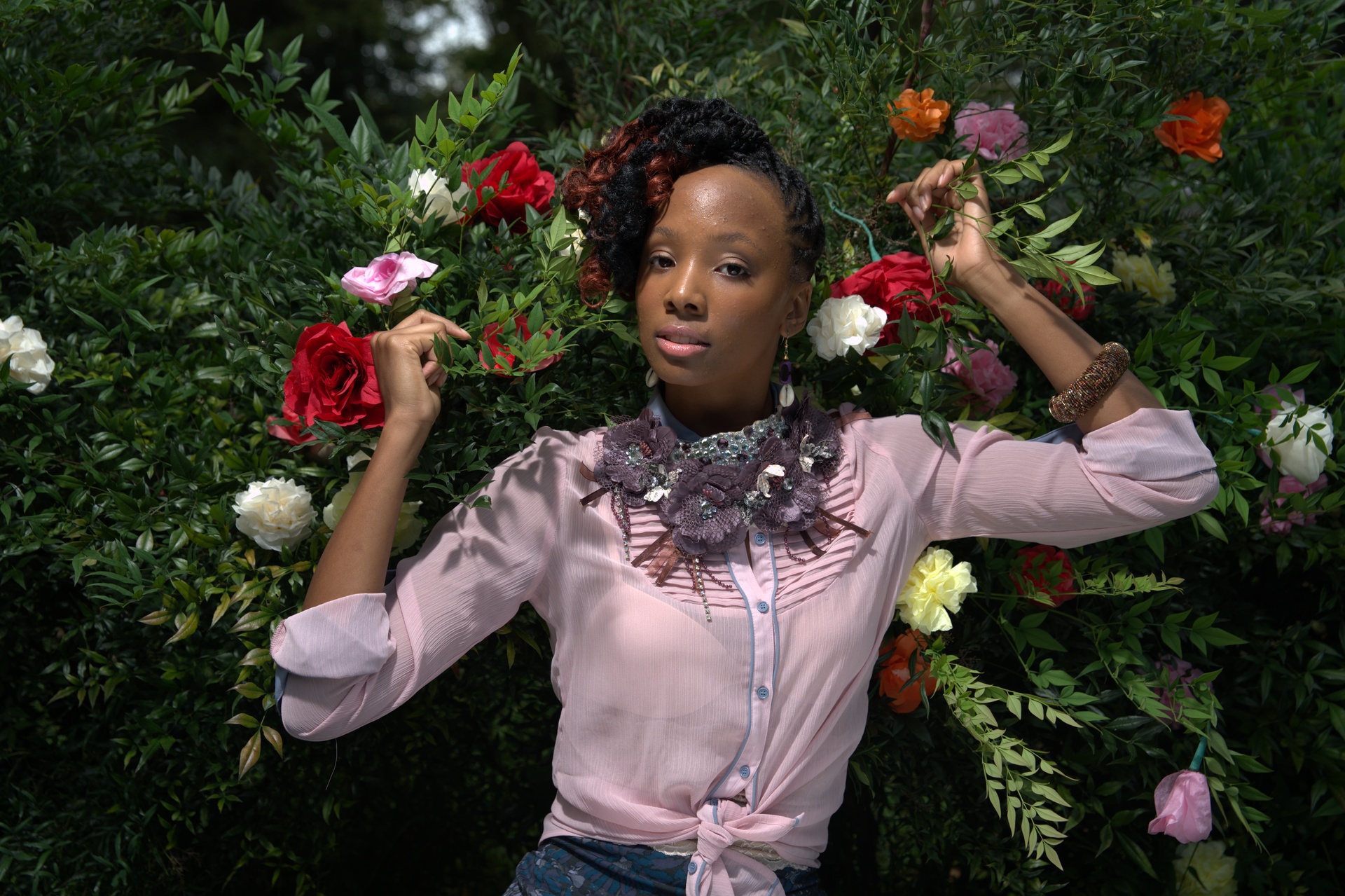}};
        \node[below=0.5mm of imgB] {Mem score: 0.752};
    \end{tikzpicture}
\end{minipage}
\hfill
\begin{minipage}{0.45\textwidth}
    \textbf{Feedback $a$:}
    \begin{enumerate}
    \item Adjust the person's posture to a standing position.
    \item Raise both arms to hold flowers above the head.
    \item Ensure the person is surrounded by lush greenery.
    \item Remove the basket and gardening tools from the scene.
    \item Adjust the lighting to create a brighter and more vibrant atmosphere.
    \end{enumerate}
\end{minipage}

\hrule
\vspace{1em}

\centering
\begin{minipage}{0.25\textwidth}
    \centering
    \begin{tikzpicture}
        \node[draw=red, line width=1pt] (imgA) {\includegraphics[width=\linewidth]{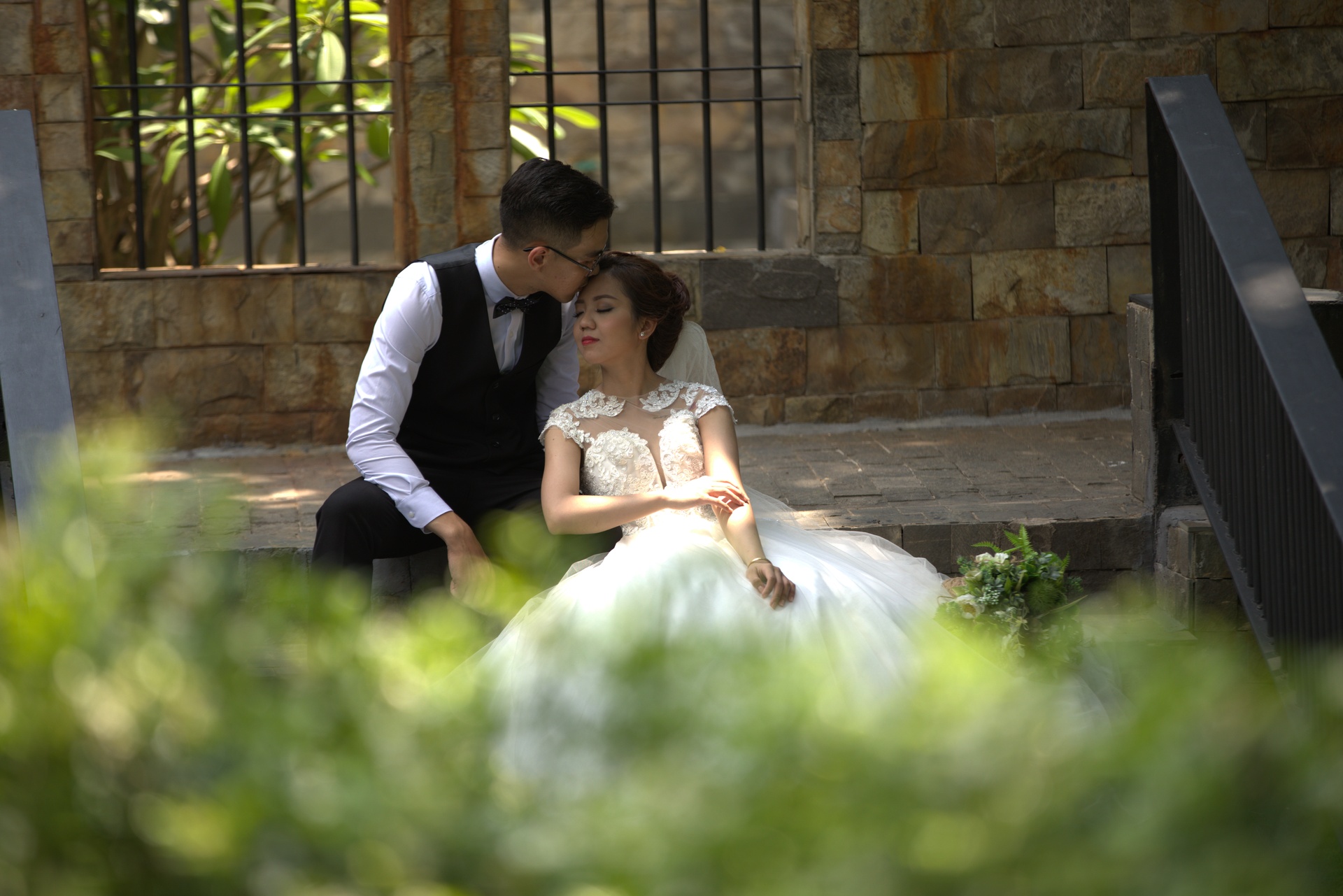}};
        \node[below=0.5mm of imgA] {Mem score: 0.807};
    \end{tikzpicture}
\end{minipage}%
\hfill
\begin{minipage}{0.13\textwidth}
    \centering
    \begin{tikzpicture}
        \node[draw=blue, line width=1pt] (imgB) {\includegraphics[width=\linewidth]{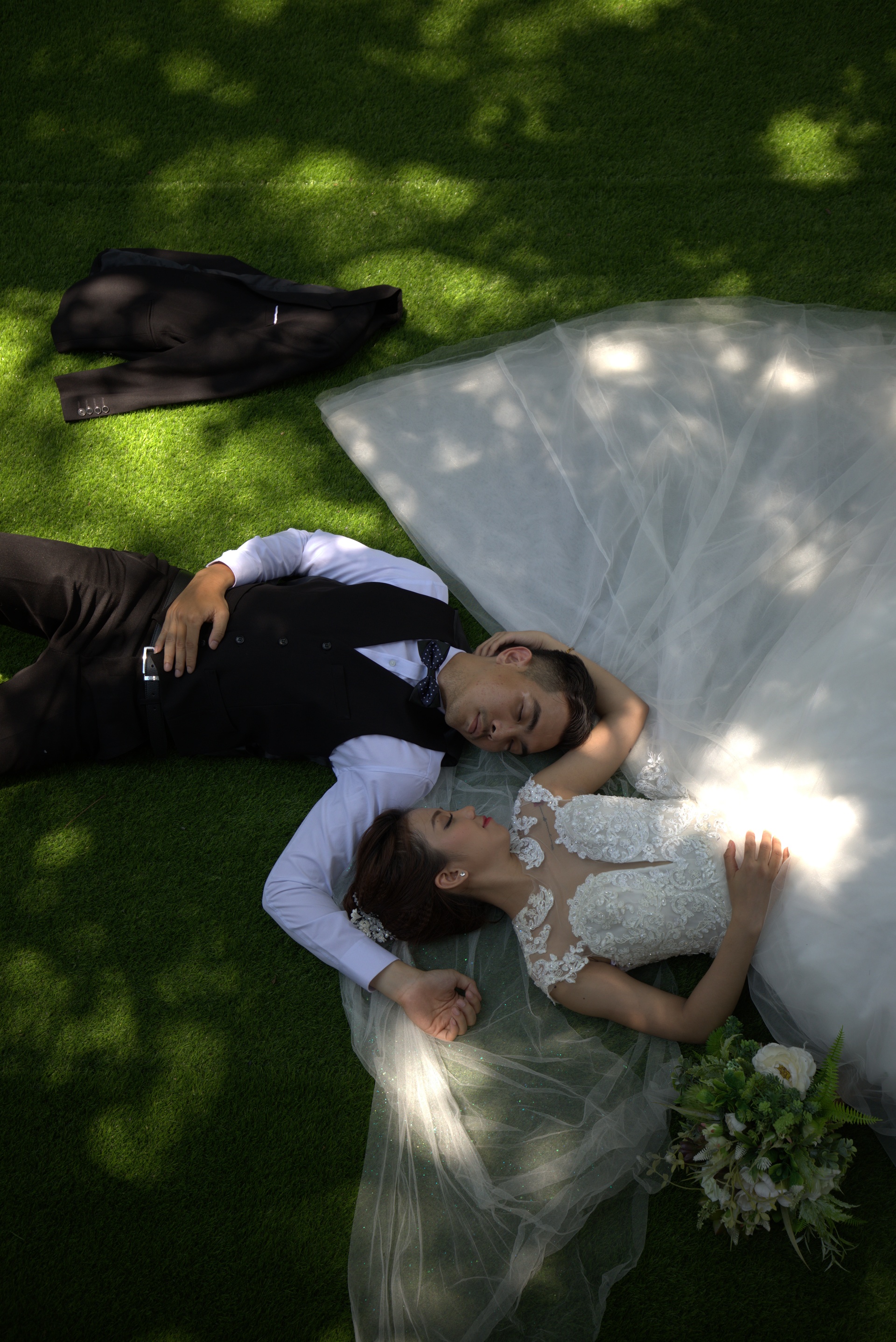}};
        \node[below=0.5mm of imgB] {Mem score: 0.892};
    \end{tikzpicture}
\end{minipage}
\hfill
\begin{minipage}{0.45\textwidth}
    \textbf{Feedback $a$:}
    \begin{enumerate}
    \item Reposition the couple from a seated to a lying down pose on a grassy surface.
    \item Adjust the angle of the shot to an overhead view.
    \item Remove the stone wall and window background, replacing it with a grassy area.
    \item Ensure the bride's dress and veil spread out naturally on the grass.
    \item Place the groom's jacket and pants neatly on the grass beside them.
    \item Position the bride's bouquet on the grass near her hand.
    \end{enumerate}
\end{minipage}

\hrule
\vspace{1em}

\centering
\begin{minipage}{0.25\textwidth}
    \centering
    \begin{tikzpicture}
        \node[draw=red, line width=1pt] (imgA) {\includegraphics[width=\linewidth]{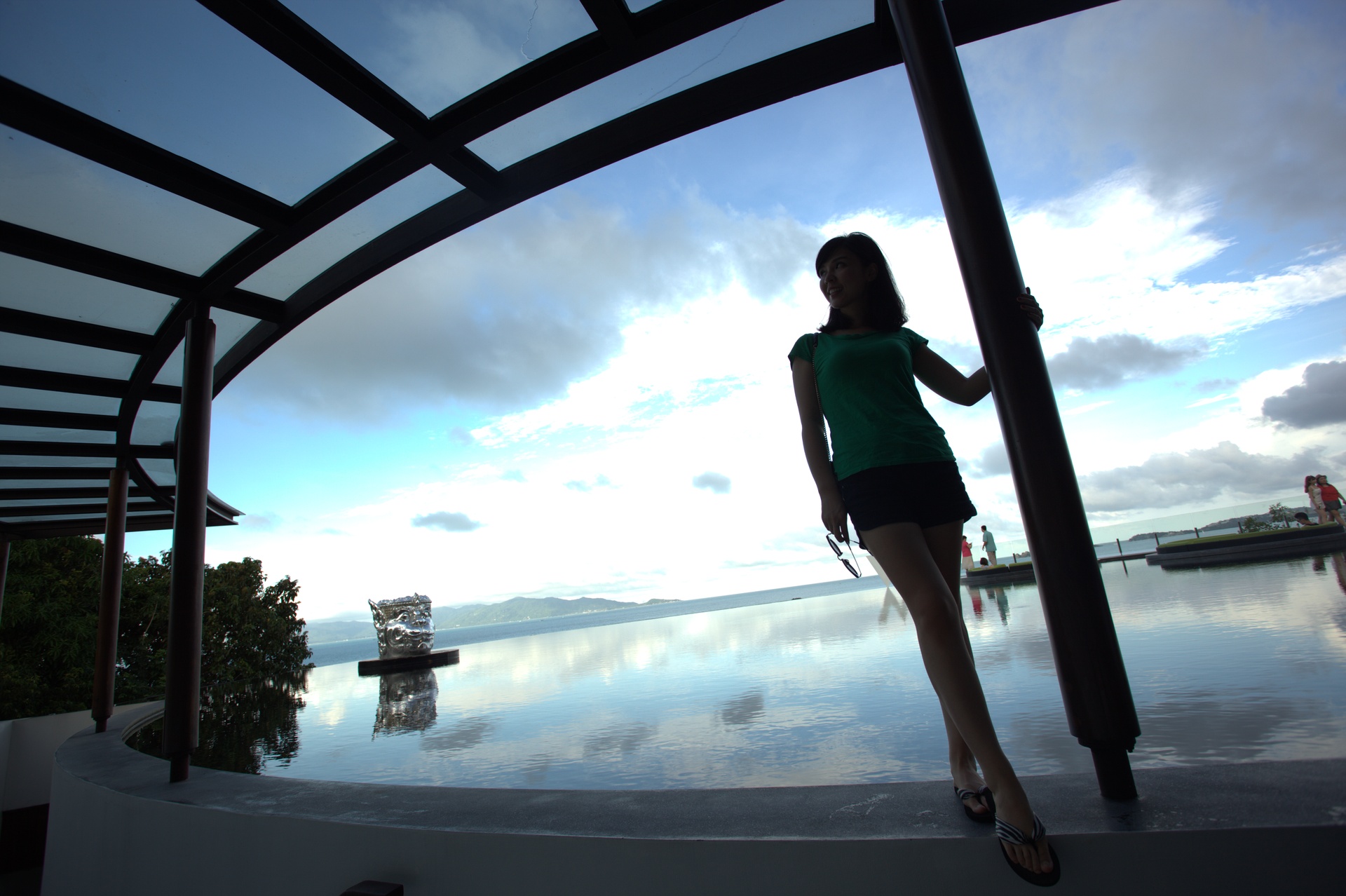}};
        \node[below=0.5mm of imgA] {Mem score: 0.749};
    \end{tikzpicture}
\end{minipage}%
\hfill
\begin{minipage}{0.25\textwidth}
    \centering
    \begin{tikzpicture}
        \node[draw=blue, line width=1pt] (imgB) {\includegraphics[width=\linewidth]{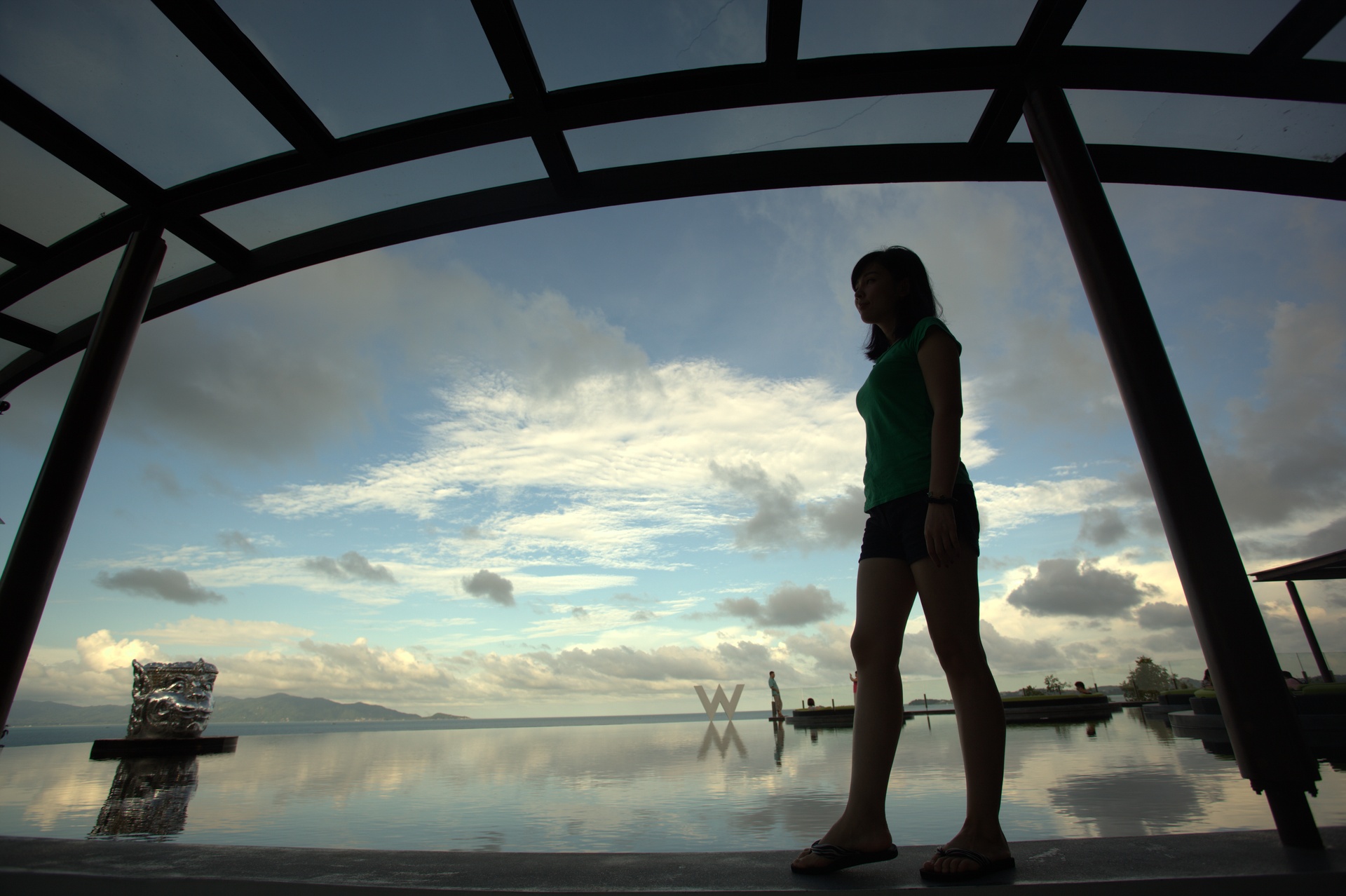}};
        \node[below=0.5mm of imgB] {Mem score: 0.829};
    \end{tikzpicture}
\end{minipage}
\hfill
\begin{minipage}{0.45\textwidth}
    \textbf{Feedback $a$:}
    \begin{enumerate}
    \item Adjust the position of the person so they are standing more centrally within the frame.
    \item Shift the perspective slightly to the right to include more of the water and the sculpture on the left.
    \item Reduce the brightness and contrast to create a softer, more subdued lighting effect.
    \item Reposition the person's arm so it is relaxed by their side, not holding onto the structure.
   \item Ensure the reflection on the water is more prominent by adjusting the angle of the light source.
    \end{enumerate}
\end{minipage}

\hrule
\vspace{1em}

\centering
\begin{minipage}{0.25\textwidth}
    \centering
    \begin{tikzpicture}
        \node[draw=red, line width=1pt] (imgA) {\includegraphics[width=\linewidth]{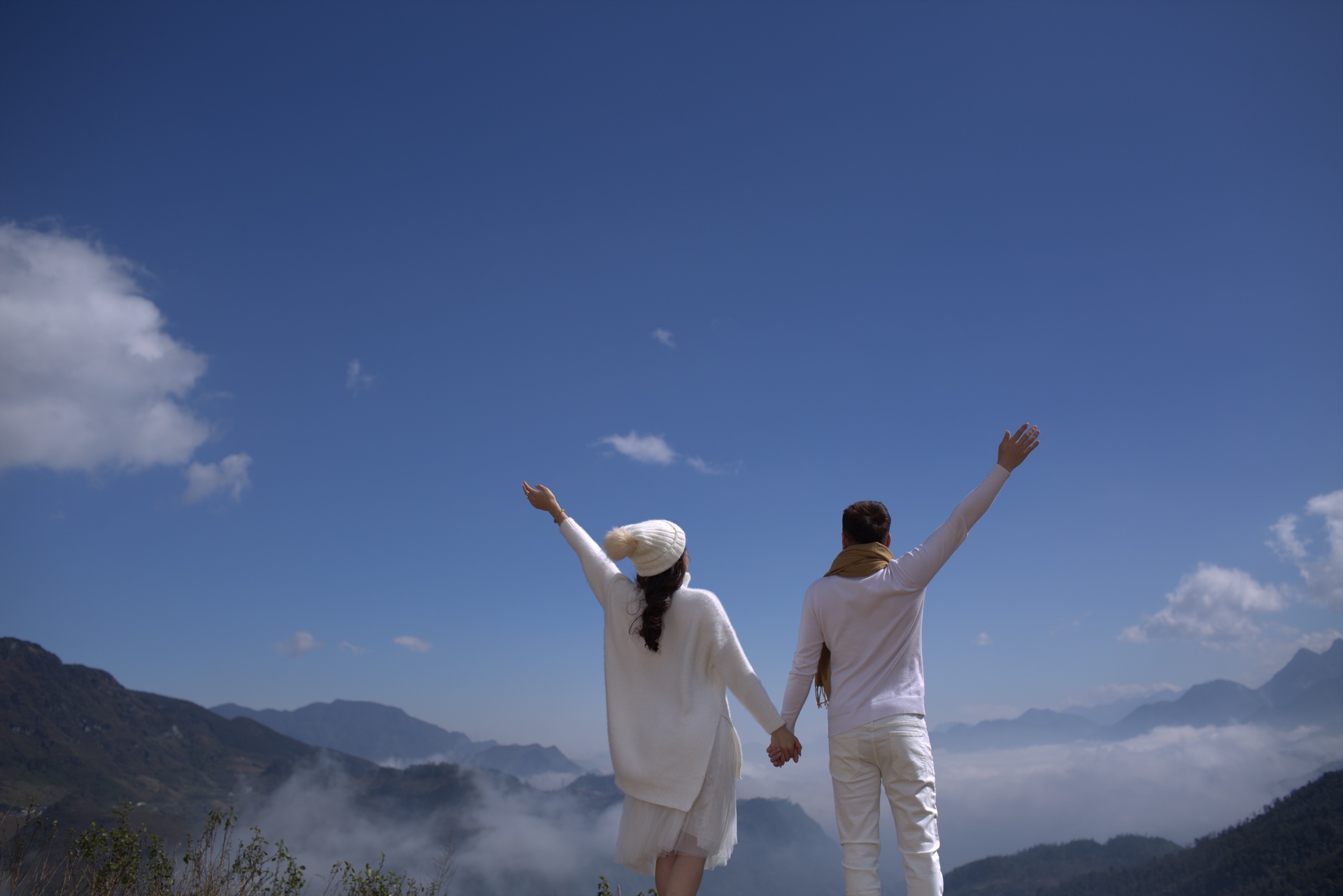}};
        \node[below=0.5mm of imgA] {Mem score: 0.798};
    \end{tikzpicture}
\end{minipage}%
\hfill
\begin{minipage}{0.25\textwidth}
    \centering
    \begin{tikzpicture}
        \node[draw=blue, line width=1pt] (imgB) {\includegraphics[width=\linewidth]{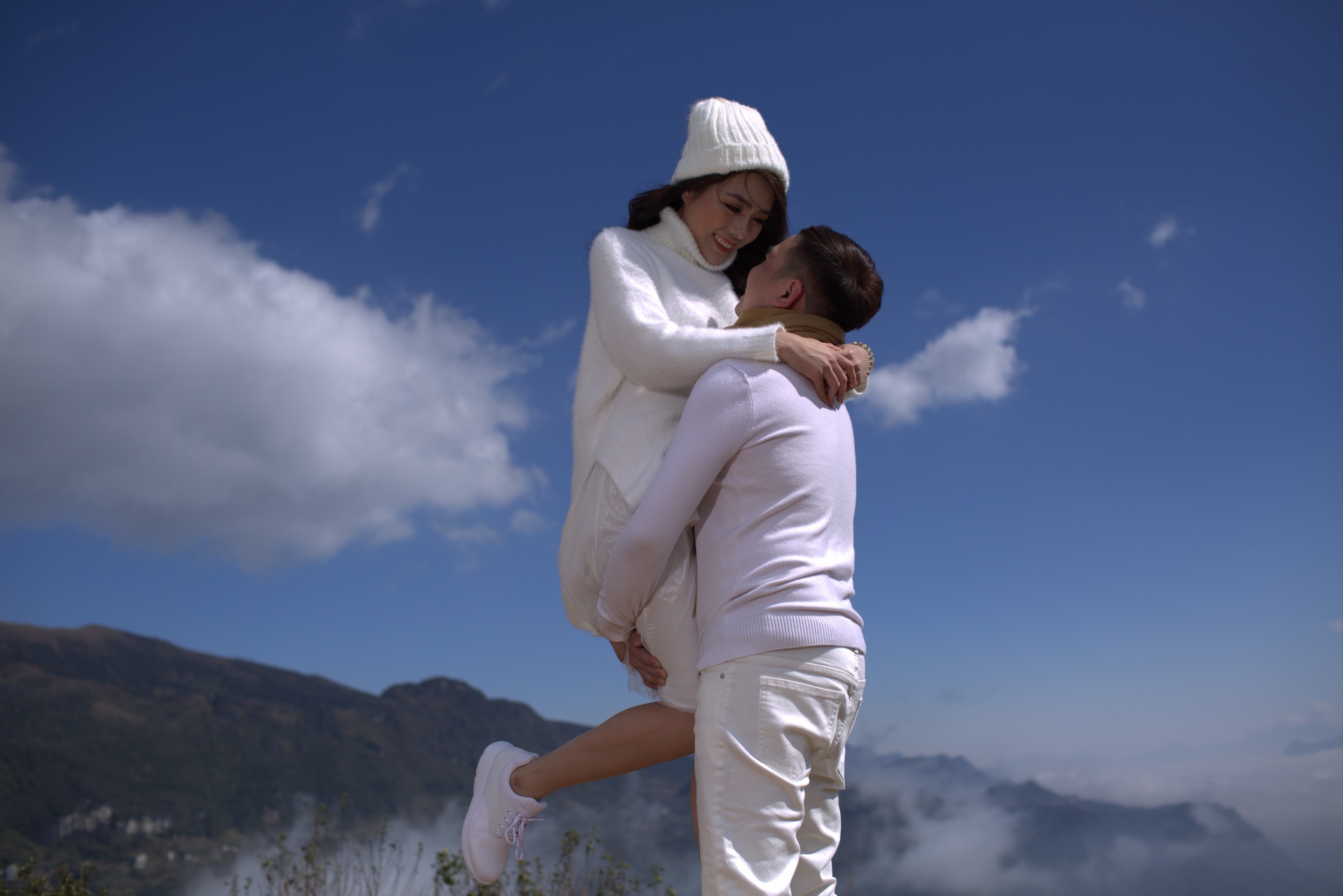}};
        \node[below=0.5mm of imgB] {Mem score: 0.990};
    \end{tikzpicture}
\end{minipage}
\hfill
\begin{minipage}{0.45\textwidth}
    \textbf{Feedback $a$:}
    \begin{enumerate}
    \item Reposition the individuals so that they are facing each other, with one person lifting the other into their arms.
    \item Adjust the arms so that the lifted person's arms are wrapped around the other person's neck.
    \item Ensure the lifted person's legs are bent at the knees and held by the other person.
    \item Shift the gaze of both individuals to look at each other affectionately.
   \item Maintain the scenic background with mountains and clouds, but adjust the angle slightly to accommodate the new pose.
    \end{enumerate}
\end{minipage}

\caption{A set of qualitative examples from MemBench.}
\label{fig:suppl_qualitatives_2}
\end{figure*}

Fig.~\ref{fig:suppl_qualitatives} and~\ref{fig:suppl_qualitatives_2} present data examples from the MemBench dataset. For each image pair, we show (from left to right) the source image (red frame), the destination image (blue frame), and the corresponding feedback generated by the multimodal model. The memorability scores assigned by the predictor $\mathcal{M}$ are shown beneath each image.

\subsection{Construction Details} 
Images within each scene are ranked using the memorability predictor $\mathcal{M}$. Pairs of least and most memorable images are then selected to construct contrastive training data. Evaluation is performed on a \textit{random held-out set of unseen scenes}, where feedback is generated starting from the scene's least memorable image.

\subsection{Image Pre-processing}
PPR10K~\cite{liang2021ppr10k} provides images in RAW format, with an average image file size of 42\,MB. To reduce storage requirements, enable efficient processing, and ensure compatibility with MLLM vision processors, we convert all RAW files to JPEG format while preserving their original aspect ratio. Conversions are performed using \texttt{rawpy} and \texttt{PIL} Python libraries, both executed with default parameters.

\subsection{Memorability Predictor} \label{sec:supp_mem_pred}
To build the memorability predictor $\mathcal{M}$, we follow the approach proposed in~\cite{zalcher2025dontjudgeclipunified} where a frozen visual feature extractor is followed by an MLP head trained for regression. The final model outputs a single continuous value in $[0,1]$ corresponding to the predicted memorability score.

\paragraph{Training.}
We train the regressor $\mathcal{M}$ on three widely used memorability datasets: LaMem~\cite{khosla2015understanding}, MemCat~\cite{goetschalckx2019memcat}, and SUN~\cite{fajtl2018amnet}. Each image is associated with a ground-truth memorability score in the range $[0,1]$. For feature extraction, we employ the vision tower of OpenCLIP~\cite{openclip_library}, specifically the \texttt{ViT-SO400M-14-SigLIP-384} model~\cite{zhai2023sigmoid} pretrained on WebLI~\cite{caron2024web}. The resulting visual embeddings (dimension 1152) serve as input to the regression head.
The MLP consists of two fully connected layers: a 256-dimensional hidden layer with a ReLU activation~\cite{agarap2018deep}, followed by a 1-dimensional linear output layer. The model is trained end-to-end only on the MLP parameters, using the mean squared error (MSE) loss. We train for 100 epochs using the Adam optimizer~\cite{kingma2017adammethodstochasticoptimization} with a learning rate of $1 \times 10^{-4}$ and weight decay set to $0.0$. Empirically, we find that using the raw (unnormalized) OpenCLIP features leads to improved Spearman’s rank correlation ~\cite{spearmanr}. A summary of the model performance is reported in Tab.~\ref{tab:supp_trained_mem_model}.

\begin{table}[ht]
\centering
   \caption{\textbf{Model $\mathcal{M}$ performance.} Comparison of memorability predictor models trained on CLIP-like embeddings. We report feature dimensionality, whether feature normalization is applied, and Spearman’s rank correlation. The \inlineColorbox{memregressor!75}{used predictor model} achieves the highest correlation among all evaluated variants. (*) is reported from \cite{zalcher2025dontjudgeclipunified}.}
    \label{tab:supp_trained_mem_model}
\resizebox{\linewidth}{!}{%
    \begin{tabular}{lccc}
        \toprule
        \textbf{Model}        & \multicolumn{1}{c}{\textbf{Fts dim}} & \multicolumn{1}{c}{\textbf{Normalization}} & \multicolumn{1}{c}{\textbf{\makecell{ Spearman \\ Rank ($\uparrow$)}}} \\
        \midrule
        \rowcolor{oraclered!35} Human*                           &      \textit{n.d.}                       &  \textit{n.d.}                                & 0.68 \\
        ViT-L-14-quickgelu                             & 768                        & \checkmark         & 0.73 \\
        ViT-L-14-quickgelu                                  & 768                        & $\times$        & 0.81 \\
        ViT-SO400M-14-SigLIP-384                                & 1152                          & \checkmark       & 0.76 \\
        \rowcolor{memregressor!75} \textbf{\benchmarkname~$\mathcal{M}$}                                & 1152                          & $\times$        & \textbf{0.82} \\
        \bottomrule
    \end{tabular}
}
\end{table}

\paragraph{Validation.}
To assess the effectiveness of our memorability predictor $\mathcal{M}$, we compare it against state-of-the-art memorability models~\cite{tai2017memnet, fajtl2018amnet, hagen2023image, si2024longtermadmemorabilityunderstanding, zalcher2025dontjudgeclipunified} on the LaMem test set (Tab.~\ref{tab:supp_comparison_mem_models}). While prior approaches are typically trained solely on LaMem, except for \cite{si2024longtermadmemorabilityunderstanding}, our model $\mathcal{M}$ leverages three datasets (LaMem, MemCat, and SUN), achieving the highest Spearman’s rank correlation among all evaluated methods.

\begin{table}[ht]
\centering
    \caption{\textbf{Comparison with state-of-the-art memorability predictors.}  
Spearman Rank correlation on the LaMem test set. Our model, trained on LaMem, MemCat, and SUN, achieves the highest correlation among all methods. (*) is reported from \cite{zalcher2025dontjudgeclipunified}.}

\label{tab:supp_comparison_mem_models}
\resizebox{0.7\linewidth}{!}{%
\begin{tabular}{lcc}
    \toprule
    \textbf{Model} & \textbf{Pretrained} & \multicolumn{1}{c}{\textbf{\makecell{Spearman\\ Rank ($\uparrow$)}}} \\
    \midrule
    MemNet* \cite{tai2017memnet} & LaMem & 0.64 \\
    AMNet* \cite{fajtl2018amnet} & LaMem & 0.67 \\
    \rowcolor{oraclered!35} Human* & \textit{n.d.} & 0.68 \\
    ViTMem* \cite{hagen2023image} & LaMem & 0.71 \\
     Henry \cite{si2024longtermadmemorabilityunderstanding} & LaMem + MemCat + SUN & 0.72 \\
    Henry \cite{si2024longtermadmemorabilityunderstanding} & LaMem & 0.74 \\
    PerceptCLIP \cite{zalcher2025dontjudgeclipunified} & LaMem & 0.74 \\
    \midrule
    \rowcolor{memregressor!75}\textbf{\benchmarkname~$\mathcal{M}$} & LaMem + MemCat + SUN & \textbf{0.82} \\
    \bottomrule
\end{tabular}
}
\end{table}

\paragraph{Limitations}.
Our automatic pipeline relies heavily on the initial ranking of images, which is determined by a memorability predictor $\mathcal{M}$. Although this dependency introduces a potential source of bias, we treat the memorability model as a well-established black-box component, consistent with prior literature~\cite{goetschalckx2019memcat, isola2011makes, newman2020multimodal, lahrache2022survey, perera2019image, zalcher2025dontjudgeclipunified, isola2011understanding, isola2013makes, bylinskii2015intrinsic, khosla2015understanding}. In this sense, our framework is agnostic to the specific predictor used: given any target scoring criterion, it may substitute an alternative ranking signal and construct systems tailored to their own objectives.

\subsection{Editing Baseline} \label{sec:img_edit_details}
For in-context image editing, we employ \textsc{Flux.1 Kontext}~\cite{labs2025flux1kontextflowmatching}, as described in Sec.~\ref{sec:membench}, via \texttt{diffusers} library~\cite{diffusers_library} from HuggingFace (open-source model version with tag  \texttt{FLUX.1-Kontext-dev}). We use the default configuration recommended by the original authors, setting the number of inference steps to 28, the guidance scale to 2.5, and fixing the seed generator to 0 to ensure reproducibility. The default aspect ratio of the model is 1:1, specifically $1024\times1024$ pixel resolution.

\subsection{Prompting and Structured Feedback} \label{sec:supp_membench_details}

\paragraph{Feedback elicitation prompt.}
To generate the full MemBench dataset, we rely on prompt $p_a$, presented in Sec.~\ref{sec:membench}. The complete prompt $p_a$ is reported below.

\begin{systemmsg}
\textbf{System:} You are an observer.
\end{systemmsg}

\begin{usermsg}
\textbf{User:}

\begin{center}
    \includegraphics[width=0.44\linewidth]{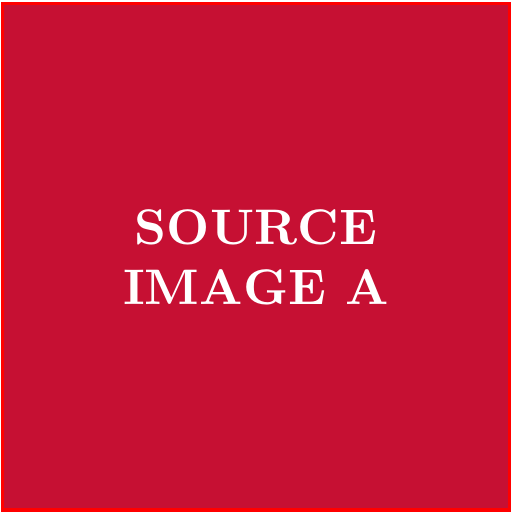}
    \hfill
    \includegraphics[width=0.44\linewidth]{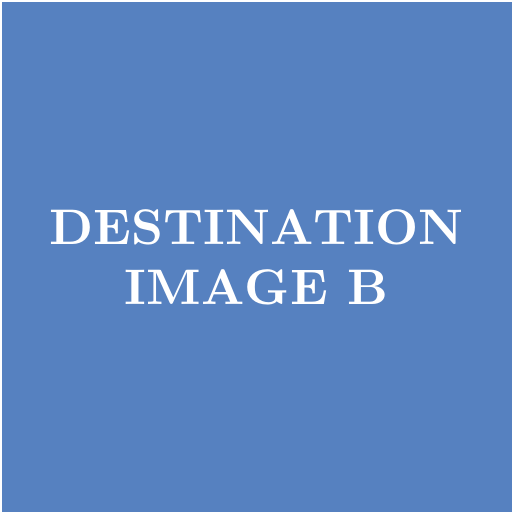}
\end{center}

Your task is to determine the actions required to transform Image A into Image B. \\
Strictly avoid both explicit and implicit references to the images when suggesting action items, and ensure that each action item is fully self-contained. \\[4pt]

Produce a structured JSON object that must include: \\
\quad-- \texttt{actions}: a list of precise and well-informative semantic actions. \\[4pt]

Respond with a valid JSON object and no explanation.
\end{usermsg}

\begin{qwencaption}
\textbf{Output:}
\begin{verbatim}
{
    "actions": [
        <#1 sub-action>,
        <#2 sub-action>,
        ...,
        <#k sub-action>
    ]
}
\end{verbatim}
\end{qwencaption}

\noindent The resulting output is then parsed as a JSON object.

During early experimentation, models frequently produced feedback that referred directly to the target image rather than describing the transformation itself (\eg{} ``\textit{Adjust the brightness to match the one in Image B}''). To address this issue, we refined the prompt to explicitly forbid image-referential phrasing and require self-contained action descriptions.

In cases where multiple source images receive the same memorability score, ties are resolved by sorting according to the filename identifier in descending order.

\paragraph{Structured, formally valid JSON outputs.}
To ensure consistently structured outputs while maintaining flexibility in the definition of output fields, we adopt the \texttt{outlines} library~\cite{willard2023efficient} for constrained decoding; compatible with the \texttt{transformers} library~\cite{wolf-etal-2020-transformers}, it allows to enforce a predefined output schema. For extracting the feedback divided into subactions, we define the following class specification:
\begin{listing}[ht] %
\begin{lstlisting}[language=Python]
class ActionListOutput(BaseModel):
    actions: List[str] = Field(
        description="A list of actions.",
        min_items=1,
        max_items=10
    )
\end{lstlisting}
\caption{Class specification to ensure valid JSON schema.}
\label{lst:supp_actionlist}
\end{listing}

This setup enforces syntactic validity, guarantees reliable parsing, and enables systematic storage of feedback samples in JSON format. The schema-based design also allows for straightforward modifications, such as adding or removing fields when extending the output format.

\subsection{Feedback Sub-actions Categorization}
To categorize the atomic sub-actions contained in each feedback instance, we employ \textsc{GPT-5 Mini}~\citep{OpenAI_2025_GPT5_System_Card} as an automatic annotator. The model is prompted with a taxonomy covering six high-level categories: \emph{Framing}, \emph{Lighting}, \emph{Posing}, \emph{Semantics}, \emph{Intent}, and \emph{Aesthetics}. The prompt used for annotation is reported below:

\begin{usermsg}
\textbf{User:}

Consider the following action that a photographer can do. Categorize the action into: \\
FRAMING -- Zoom/Crop/Reframe, Angle and Viewpoint, Balance and Symmetry\\
LIGHTING -- Lighting direction/strength/temperature, Exposure adjustment, Shadows control\\
POSING -- Pose adjustments, Facial expressions, Subjects interaction, Clothes\\
SEMANTICS -- Add/remove objects or people, Change background, Include contextual cues\\
INTENT -- Change narrative emphasis, Mood and atmosphere\\
AESTHETICS -- Color grading/filters, Contrast/sharpness, Blur and focus\\[3pt]
Here the action: \texttt{<input action>}
\end{usermsg}

This setup guarantees consistent labeling across all sub-actions, enabling downstream analysis of category frequencies and co-occurrence patterns.

\section{\methodname Additional  Details} \label{suppl:membench_tech_details}

MemCoach is a training-free method that applies activation steering to modulate the internal representations of a multimodal model. All experiments were conducted using the PyTorch framework on a single NVIDIA A100 GPU (64 GB). 

Below, we provide the implementation details for generating and injecting the steering vector, complementing the description in Sec.~\ref{sec:memolift}. 

\paragraph{Extraction stage.}
We target the residual module within a specific language Transformer block of the multimodal backbone (\eg, layer $l = 55$ in our best-performing configuration). For a chosen layer~$l$, we register a forward hook to capture the activation tensor $h^{(l)}$ for each input sequence~$i$. Each input sequence, as defined in Eq.~\ref{eq:input_steering_vects}, is first tokenized into input IDs. For each input $i$, we compute the mean over the sequence dimension to obtain a single activation representation. No normalization or additional post-processing is applied. Since we operate in batches, we record the starting index of padding tokens and exclude padded positions from the mean computation. No generation is performed during this stage: the model is only run in forward mode.  
The \gradienttext{memorability steering vector} $\mathbf{r}^{(l)}$ is computed as described in Sec.~\ref{sec:memolift}.

\paragraph{Inference stage.}
Inference requires selecting a steering coefficient~$\alpha$ and the target layer~$l$. During the model’s forward pass, the vector~$\mathbf{r}^{(l)}$ is injected, scaled by~$\alpha$, at the specified layer, following Eq.~\ref{eq:steered-activation}. Injection is applied uniformly along the sequence dimension. The forward computation then proceeds as usual, but with altered activation patterns at layer~$l$, steering the model toward memorability-aware behaviour.

\paragraph{Different models configuration.}
Tab.~\ref{tab:supp_comparison_layer_coeff} reports the optimal layer index $l$ and steering coefficient $\alpha$ identified for the four open-source models we evaluate. Hyperparameters are optimized on a held-out split of the training set.

\begin{table}[ht]
    \centering
    \caption{\textbf{Optimal steering configuration across models.}  
Best-performing layer index $l$ and steering coefficient $\alpha$ obtained via hyperparameter tuning for each evaluated open-source multimodal model. We also report the language-model depth (LM Depth) and the corresponding IR score.}
    \label{tab:supp_comparison_layer_coeff}
    \resizebox{\linewidth}{!}{%
    \begin{tabular}{lcccc}
        \toprule
        \textbf{Model} & \textbf{Layer} & \textbf{Coefficient} & \textbf{LM Depth} & \textbf{IR} \\
        \midrule
        \textit{\textbf{\methodname-\textsc{Idefics}}}  & 13 & 30  & 32 & 0.75 \\
        \textit{\textbf{\methodname-\textsc{LLaVA}}}    & 20 & 143 & 28 & 0.73 \\
        \textit{\textbf{\methodname-\textsc{Qwen}}}     & 12 & 26  & 28 & 0.74 \\
        \textit{\textbf{\methodname-\textsc{InternVL}}} & 12 & 55  & 36 & 0.80 \\
        \bottomrule
    \end{tabular}
    }
\end{table}

\paragraph{Implication of optimizing memorability.} 
Enhancing memorability raises ethical concerns, including potential manipulation, undue influence on viewer perception, and the risk of homogenizing visual expression by favoring conventional cues over diversity. At the same time, in assistive and controlled contexts, \eg education, creative exercises, or personal photography coaching, memorability optimization can enhance communication, reinforce learning recall, and provide actionable guidance without overriding individual intent. Balancing these risks and benefits is essential, emphasizing the need for transparency, user agency, and context-aware application to ensure that memorability interventions remain supportive rather than prescriptive.

\section{User Studies} \label{sec:human-studies-appendix}
The following experiments are preliminary user studies designed to evaluate the effectiveness of \methodname. 
They aim to provide early validation and insights on the effect of \methodname on human memorability (\cref{sec:supp:user:human-mem}), evaluate the effectiveness of the approach for real-life guidance (\cref{sec:supp:user:human-in-the-loop}), and probe the quality of feedback according to users (\cref{sec:supp:user:quality}).

\subsection{Human Memorability Alignment}
\label{sec:supp:user:human-mem}
To assess how images from different settings drive human memorability, we conduct a memorability experiment with 47 valid users (\textit{avg.} 15.6 annotations/image), following previous  work~\citep{goetschalckx2019ganalyze}. 

\paragraph{Experiment setup details.} 
Participants completed a continuous recognition (repeat-detection) visual memory task in which they viewed a stream of images and pressed the space bar whenever they detected a repeat of an image previously shown within the same session. Each session consisted of 150 images presented sequentially, with each image displayed for 600 ms followed by an 800 ms blank interstimulus interval. The sequence contained 40 target pairs, where each target image appeared once and was repeated after 22-93 intervening images. An additional 10 vigilance pairs were included as attention checks, in which the repeat occurred after 1-4 intervening images. The remaining 50 images were fillers presented only once to maintain spacing and reduce predictability of repeats. Participants could respond at any time during the 1400 ms trial window (image plus blank interval). The same task protocol was used across three stimulus variants, \textit{i.e.} \methodname, \internvl~, and source images $x_s$, with identical timing and sequence structure; only the image sources differed. Sequence order was deterministic and reproducible, using a fixed seed. Participants were instructed not to repeat the game after completing one. If participants missed more than 50\% of the vigilance repeats in a run, user results were excluded from the analyses.

\begin{figure}[ht]
    \centering
    \includegraphics[width=.55\linewidth]{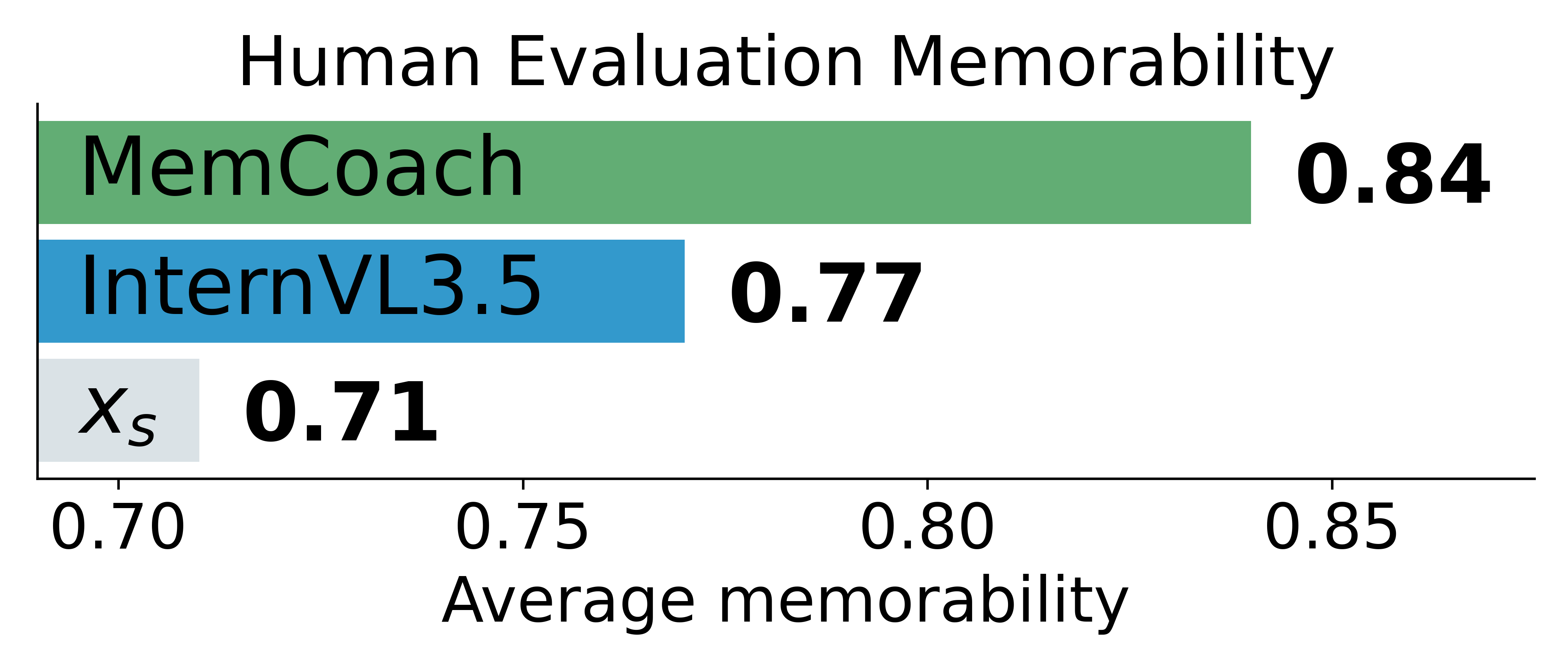}
    \caption{\textbf{Human memory performance} across three different image settings.}
    \label{fig:human_perf}
\end{figure}

\paragraph{Results.} Consistent with the editing experiments, Figure \ref{fig:human_perf} shows both \internvl~ and \methodname increase the average memorability \textit{wrt} source images $x_s$, with \methodname achieving a large margin of improvement. The gap between methods indicates that \methodname more effectively shifts images toward higher memorability regimes, while \internvl~ provides only moderate improvements. Overall, results highlight the importance of explicitly injecting memorability-aware signals to achieve stronger memorability outcomes.

\subsection{Human-in-the-loop Evaluation}
\label{sec:supp:user:human-in-the-loop}
To evaluate real-world effectiveness, we conduct a preliminary human-in-the-loop evaluation measuring whether users can successfully follow the generated feedback to produce more memorable images. 

\paragraph{Experiment setup details.}  We implemented a mobile app allowing a user to use \methodname in real-life scenarios. The app presents a live camera viewfinder and allows users to point their phone at any scene of their choice. Upon capturing a frame, users could request either a memorability score alone or a memorability score accompanied by actionable textual feedback generated by \methodname. 
\begin{wrapfigure}{l}{0.35\linewidth}
    \includegraphics[width=\linewidth]{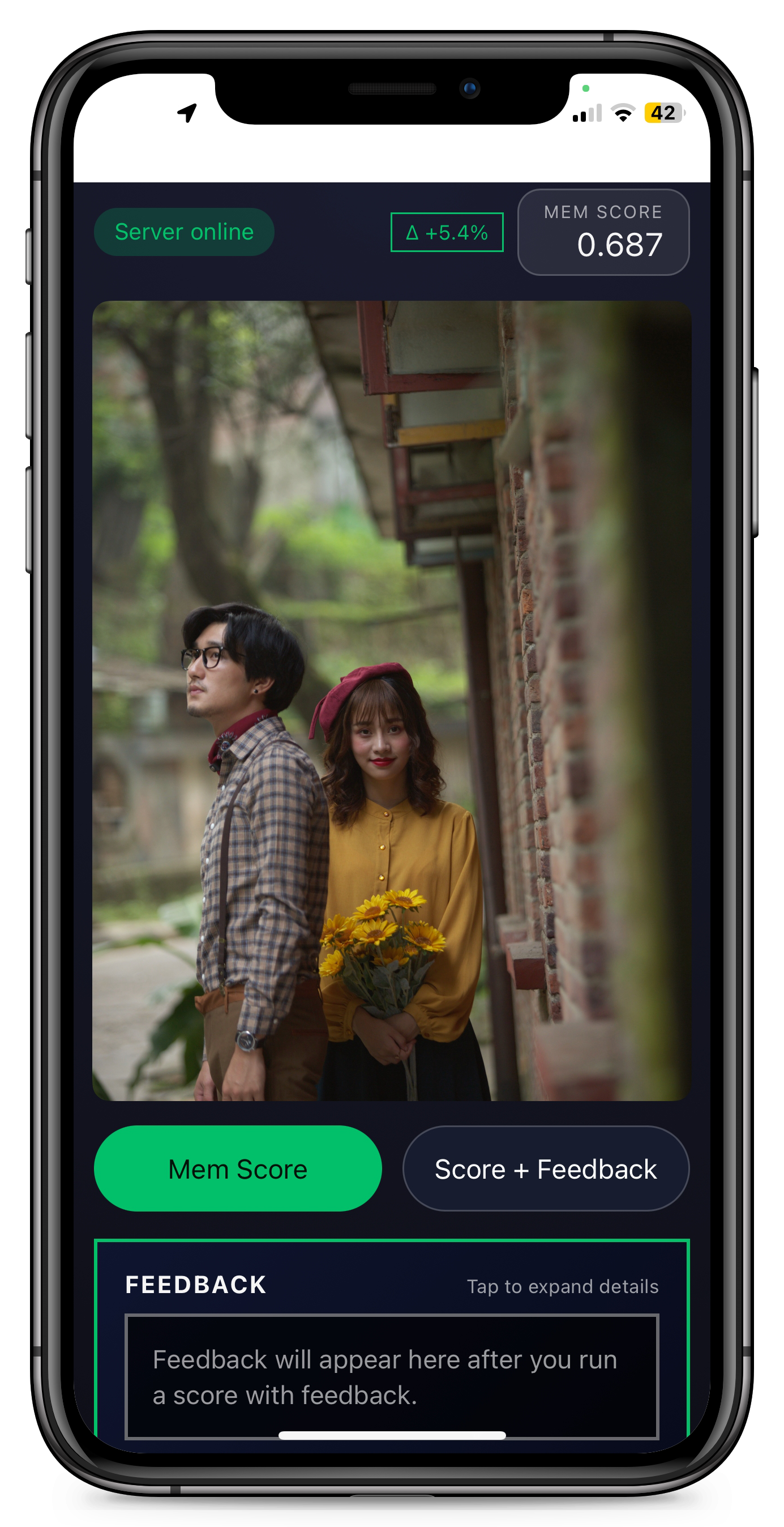}
\end{wrapfigure}
The feedback is displayed directly on screen as a natural-language suggestion, guiding the user toward a more memorable composition. Users are free to implement the feedback by adjusting their framing or scene and re-capturing the image, observing how the memorability score changed between attempts. When the feedback pertained to the objects or subjects of the scene, the phone was kept as stable as possible between captures to isolate the effect of compositional changes. Each submitted image, together with its predicted score and generated feedback, was logged for offline analysis. The app requires no installation and runs entirely in the browser, served over a secure HTTPS connection via a public tunnel to ensure accessibility across devices and operating systems.

\paragraph{Results.} We collected 27 scenes and evaluated them using the memorability predictor $\mathcal{M}$. Despite the domain shift, \methodname consistently improves memorability, achieving an IR of $0.52$ and a relative gain (RM) of $+4.9\%$. These improvements reflect both the frequency and magnitude of successful memorability enhancements, indicating that the generated feedback effectively guides actionable changes even in previously unseen scenarios. Qualitative inspection confirms that suggestions focus on semantically meaningful adjustments, such as subject positioning, gaze direction, and interaction cues, rather than superficial edits. Overall, these results demonstrate the potential of human-in-the-loop memorability optimization and motivate future exploration of larger-scale, user-centered studies and strategies for robust real-world deployment.

\subsection{Feedback Quality Evaluation}
\label{sec:supp:user:quality}
To understand the effectiveness of memorability feedback beyond score improvements, we conducted a human study with 28 participants (381 annotations), rating the generated feedback from MemCoach on a 1–5 Likert scale along three dimensions: \textit{Clearness} (clarity of steps), \textit{Relevance} (scene-specificity), and \textit{Feasibility} (realism of applying it). 

\paragraph{Experiment setup details.} Participants completed a feedback-rating task in which they viewed a source image together with feedback and provided three scalar judgments about the feedback. For each item, participants were asked to rate:
\begin{enumerate}[(i)]
    \item \textit{Clearness}. ``How clear are the steps needed to change the photo?'' (1: Not at all, 5: Very clear);
    \item \textit{Relevance}. ``How specific is the advice for this scene?'' (1: Very general, 5: Very specific);
    \item \textit{Feasibility}. ``How realistic and sensible is this feedback to apply in real-world conditions (e.g., given physical constraints, tools, and context)?'' (1: Not realistic or sensible, 5: Completely realistic and sensible).
\end{enumerate} 
Each item yielded three numeric ratings, and items were presented in a randomized order with identical instructions and scale anchors across conditions, ensuring consistent evaluation of clarity, scene-specificity, and real-world actionability.

\paragraph{Results.}
A summary of the experiment performance is reported in Tab. \ref{tab:feedback-quality-eval}. \benchmarkname Oracle achieves consistently high scores across all dimensions, confirming the advantage of access to memorability-aware privileged information. \methodname maintains strong clearness while improving feasibility, indicating that its suggestions are generally easier to interpret and implement in practice. This gain, however, comes with a slight reduction in scene-specificity, suggesting a tendency toward more generic but broadly applicable guidance. Overall, the results highlight a trade-off between precision and practicality, where \methodname favors actionable and reliable feedback over highly tailored but less consistently executable suggestions.

\begin{table}[ht]
\centering
    \caption{\textbf{Feedback quality results.}}

    \label{tab:feedback-quality-eval}
\resizebox{0.7\linewidth}{!}{%
\begin{tabular}{l ccc}
    \toprule
    \textbf{Model} & \textit{Clearness} & \textit{Relevance} & \textit{Feasibility} \\ 
    \midrule
    Oracle  & 4.19 & 4.30  & 4.11 \\
    \methodname & 3.96 & 3.76 & 4.32 \\   
    \bottomrule
\end{tabular}
}
\end{table}

\section{Additional Analyses} \label{sec:add-analysis-appendix}

\subsection{Positive and Failure Cases} 
To analyze the factors underlying successful and unsuccessful memorability feedback, we designed a structured annotation pipeline based on an LLM-as-judge approach that classifies each sample according to a fixed taxonomy.

\paragraph{Experiment setup details.}
For each sample, the judge receives three inputs: the original source image, the generated feedback text, and the measured memorability outcome (Improved or Worsened) as determined by our memorability predictor $\mathcal{M}$. Based on the outcome, one of two dedicated prompt templates is selected. Both templates frame the model as an expert annotator performing classification rather than creative analysis, but present mutually exclusive category sets tailored to the direction of change. For improved samples, the judge selects among five improvement categories: (i) Posing / Body Configuration, (ii) Framing / Composition, (iii) Lighting / Visibility, (iv) Semantic Clarity, and (v) Emotional or Social Salience. For worsened samples, it selects among five failure categories: (i) Template Over-Normalization, (ii) Distinctiveness Suppression, (iii) Feasibility / Actionability Failure, (iv) Attention Dilution, and (v) Perceptual Degradation. In both cases, the model is required to first produce a one-sentence justification grounded strictly in the visible image content and the provided feedback, serving as a reasoning step before the final category assignment, followed by a single primary category, with explicit prohibitions against introducing new categories, referencing memorability scores, or speculating beyond the observable evidence. Annotations have been generated using \modelGPTmini~ model.

\paragraph{Results.}
Positive effects are driven primarily by posing (74.9\%), which plays the dominant role, followed by emotional saliency (23.11\%), and to a much smaller extent by framing (1.59\%). This distribution indicates that improvements are largely attributable to how subjects are positioned and the emotional cues conveyed, with only marginal influence from compositional framing.
Conversely, failures are predominantly caused by over-normalization (76.19\%), which emerges as the principal limiting factor, along with distinctiveness suppression (22.22\%), which further reduces the effectiveness of the outcome. Additional failure modes contribute only marginally, including perceptual degradation (1.59\%), while instances of attention dilution or action infeasibility occur only rarely and have a negligible overall impact.

\subsection{Editing Instruction-Following} 
To decouple feedback quality from the editor’s instruction-following, we compare memorability changes from (i) human-in-the-loop ground truth (\cref{sec:supp:user:human-in-the-loop}) and (ii) edited images using the same feedback. Similar performance is noted ($0.52$ IR, $+4.9$\% RM ground truth \textit{vs} $0.55$ IR, $+2.19$\% RM edited) and the moderate correlation ($\rho$ = 0.51) between destination image memorabilities confirms the editor as a valid proxy for automated evaluation.

\subsection{Leveraging Predictor Biases}
To mitigate shortcut learning, we ran a cross-predictor experiment using different memorability predictors for steering, \textit{i.e.} ViTMem (VM) \citep{hagen2023image}, and evaluation (our memorability predictor $\mathcal{M}$, MB). As shown in Tab.~\ref{tab:main-comparison-rebuttal}, using VM alone or the cross-predictor setup (VM $\rightarrow$ MB) consistently confirms the effectiveness of \methodname, indicating that the observed improvements are not an artifact of a single predictor. These results demonstrate that \methodname’s feedback remains robust across varying evaluation criteria, effectively enhancing memorability regardless of the specific predictor used, and mitigating concerns of shortcut learning or predictor-specific bias.

\begin{table}[ht]
\centering
    \caption{\textbf{Performance of our framework on different settings}. We report the results using different editing models for the evaluation and different memorability predictors.}
\resizebox{0.95\linewidth}{!}{%
    \begin{tabular}{l ccc cccc}
    \toprule
    \multirow{3}{*}{\textbf{Model}} &
    \multicolumn{2}{c}{\textbf{Edit Model}} &
    \multicolumn{2}{c}{\textbf{Mem Predictor}} \\
    \cmidrule(lr){2-3} \cmidrule(lr){4-5}
     &
      \textbf{Qwen-IE} &
      \textbf{FLUX.2-k} &
      \textbf{VM} &
      \textbf{VM $\rightarrow$ MB} \\
    \midrule
        \rowcolor{gray!10}\textit{Edit model} & 0.69 & 0.68  & 0.64  & 0.69 \\
    \midrule
    \rowcolor{zsblue!19}\textit{Zero-shot baselines} & & & & \\
        \gpt  & 0.78 & 0.80  & 0.73  & 0.76 \\
        \llava  & 0.59 & 0.68  & 0.76  & 0.71 \\
        \idefics  & 0.69 & 0.68  & 0.73  & 0.73 \\
        \qwen  & 0.54 & 0.61  & 0.73  & 0.69 \\
        \internvl  & 0.78 & 0.74  & 0.68  & 0.73 \\
     \midrule
     \rowcolor{methodgreen!50}\textit{MemCoach} & & & &\\
        \textit{\textbf{\methodname-\textsc{LLaVA}}}  & 0.80 & 0.73  & 0.69  & 0.74 \\
        \textit{\textbf{\methodname-\textsc{Idefics}}}  & 0.85 & 0.74  & 0.77  & 0.76 \\
        \textit{\textbf{\methodname-\textsc{Qwen}}}  & 0.82 & 0.76  & 0.77  & 0.81 \\
        \textit{\textbf{\methodname-\textsc{InternVL}}}  & \textbf{0.88} & \textbf{0.83}  & \textbf{0.83} &\textbf{ 0.82} \\
    \bottomrule

    \end{tabular}
}
\label{tab:main-comparison-rebuttal}
\end{table}
\subsection{Multiple Editing Models} 

In Tab.~\ref{tab:main-comparison-rebuttal}, we evaluate \methodname across multiple editing backbones, including Qwen-Image Edit\footnote{Model card: \url{https://huggingface.co/lightx2v/Qwen-Image-Lightning}. Weight name: \texttt{Qwen-Image-Edit-2509-Lightning-8steps-V1.0-fp32}}~\citep{wu2025qwenimage} and FLUX.2-klein\footnote{Model card: \url{https://huggingface.co/black-forest-labs/FLUX.2-klein-9B}}~\citep{blackforest_flux2_klein_2026}. Across all editors, \methodname consistently increases both the frequency and magnitude of memorability improvements compared to baseline zero-shot and default feedback. Gains are robust to variations in model architecture and editing style, indicating that the proposed approach generalizes across different latent spaces and editing mechanisms. Qualitative inspection confirms that \methodname directs edits toward semantically meaningful transformations rather than generic low-level changes, producing feedback that is both actionable and visually effective. These results reinforce that the benefit of memorability-aware guidance is model-agnostic and not confined to a specific editing pipeline.

\subsection{Generalization}
As a first study, we focus on human-centric images, given the strong influence of human presence and attributes on image memorability. To assess how well this setting generalizes beyond such content, we conduct preliminary experiments on non-human images using the same editing proxy pipeline introduced in the main paper. Results on objects and landmarks from the Yo’LLaVA dataset~\citep{yollava} indicate that \methodname performs on par with \qwen{} ($0.79$ IR), \textit{i.e.}, with no degradation in performance relative to the human domain. Exploring a broader extension beyond human-centric images is left for future work.

\newpage

\end{document}